\documentclass[11pt]{article}
\pdfoutput=1
\usepackage[]{ACL2023}

\usepackage{times}
\usepackage{latexsym}

\usepackage[T1]{fontenc}
\usepackage[utf8]{inputenc}

\usepackage{microtype}

\usepackage{inconsolata}

\usepackage{graphicx}
\usepackage{amsmath}
\newcommand\cincludegraphics[2][]{\raisebox{-0.3\height}{\includegraphics[#1]{#2}}}
\usepackage{booktabs}
\usepackage{colortbl}
\usepackage{xurl}
\usepackage{enumitem}
\usepackage{amssymb}
\usepackage{multirow}
\usepackage{array}
\usepackage{makecell}
\usepackage{tabularx}
\usepackage{caption}
\usepackage{subcaption}
\usepackage{fdsymbol}
\usepackage{wasysym}
\usepackage{xparse}

\usepackage{pifont}
\newcommand{\cmark}{\ding{51}}%
\newcommand{\xmark}{\ding{55}}%
\usepackage[normalem]{ulem}
\useunder{\uline}{\ul}{}
\title{A Multimodal Analysis of Influencer Content on Twitter}

\author{
    {\bf Danae S\'{a}nchez Villegas$^\alpha$} \quad {\bf Catalina Goanta$^\beta$} \quad {\bf Nikolaos Aletras$^\alpha$}\\
    $^\alpha$ Computer Science Department, University of Sheffield, UK\\
    $^\beta$ Utrecht University\\
    {\small
    {\tt \{dsanchezvillegas1, n.aletras\}@sheffield.ac.uk}}\\
    {\small
    {\tt  e.c.goanta@uu.nl}}%, {\tt n.aletras@sheffield.ac.uk}}
}

\begin{document}
\maketitle

\begin{abstract}
     Influencer marketing involves a wide range of strategies in which brands collaborate with popular content creators (i.e., influencers) to leverage their reach, trust, and impact on their audience to promote and endorse products or services. Because followers of influencers are more likely to buy a product after receiving an authentic product endorsement rather than an explicit direct product promotion, the line between personal opinions and commercial content promotion is frequently blurred. This makes automatic detection of regulatory compliance breaches related to influencer advertising (e.g., misleading advertising or hidden sponsorships) particularly difficult. In this work, we (1) introduce a new Twitter (now X) dataset consisting of $15,998$ influencer posts mapped into commercial and non-commercial categories for assisting in the automatic detection of commercial influencer content; (2) experiment with an extensive set of predictive models that combine text and visual information showing that our proposed cross-attention approach outperforms state-of-the-art multimodal models; and (3) conduct a thorough analysis of strengths and limitations of our models. We show that multimodal modeling is useful for identifying commercial posts, reducing the amount of false positives, and capturing relevant context that aids in the discovery of undisclosed commercial posts.\footnote{Code and data are available at \url{https://github.com/danaesavi/micd-influencer-content-twitter}} 
\end{abstract}

\section{Introduction}
\label{intro}

Social media influencers are content creators who have established credibility in a specific domain (e.g., fitness, technology), are sometimes followed by a large number of accounts and can impact the buying decisions of their followers \cite{keller2003influentials, brown2008influencer, nandagiri2018impact,lee2022consumers}. Influencer marketing (i.e., promoted content via influencer posts in social media) has gained popularity as an alternative to traditional advertising (e.g., magazines, television, billboards) and mainstream digital marketing such as pop-up and platform ads \citep{Leerssen2019,nandagiri2018impact,lou2019investigating,jarrar2020effectiveness,fang2022using} for reaching a larger and more targeted audience \citep{gross2018big}.%\footnote{Industry's overall estimated market size increase by at least 50\% each year since 2016.}%\footnote{\url{https://influencermarketinghub.com/influencer-marketing-benchmark-report-2021/}} 

%%%%%%%%%%%%%%%%%%%%% Sample Tweets %%%%%%%%%%%%%%%%%%%%%%%%%%%%%%
% For a truly beautiful and delicate summer fragrance you have to try 
% @ShayandBlue
% 's newest scent - White Peaches #BoutiquePerfumery #Fragrance 
% undiclosed 1126179120361156608
\renewcommand*{\arraystretch}{1.2}
\begin{figure}[!t]
    \footnotesize
    \centering
    %\small
    \begin{tabular}{cl}
   % \hline
        \cincludegraphics[scale=0.09]{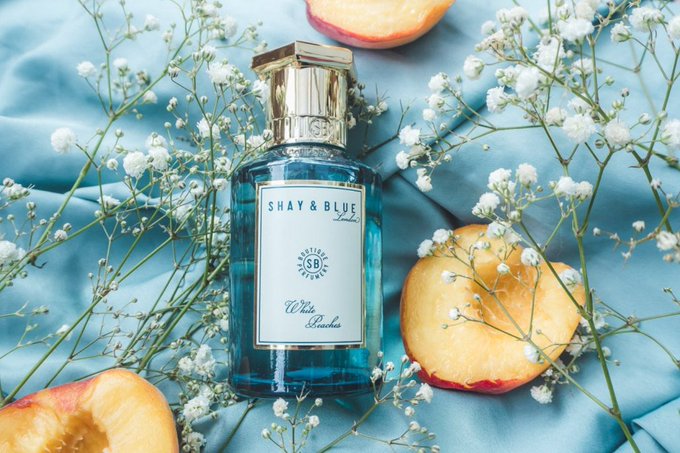}&
        \begin{tabular}[c]{@{}l@{}}\textbf{Commercial}: For a truly beautiful and\\ delicate summer fragrance you have to\\ try @USER's newest scent.\\\\ \end{tabular} \\ 
         \cincludegraphics[scale=0.04]{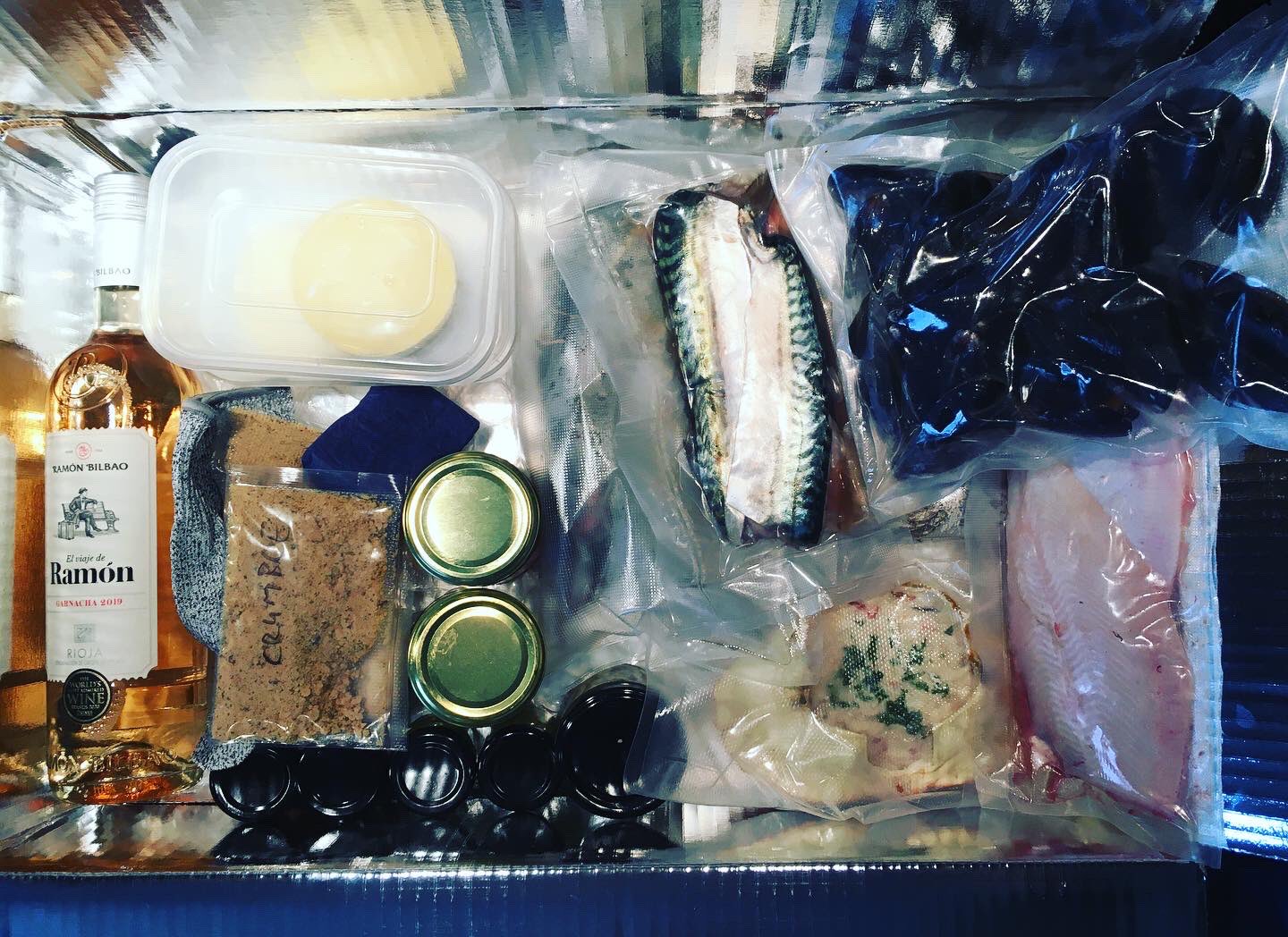} & \begin{tabular}[c]{@{}l@{}}\textbf{Non-commercial:} So that’s tonight’s\\ dinner, tomorrow’s lunch, dinner \& \\inbetweenies sorted. \\ \\ %Thanks to @USER\\ for today’s amazing delivery. \\ \\
        \end{tabular}
        %\\ \hline
    \end{tabular}

    \caption{\emph{Commercial} and \emph{non-commercial} tweets in our dataset. %The distinction between \emph{commercial} and \emph{non-commercial} posts is frequently uncertain. 
    }
    \label{fig:sampleLabels}
\end{figure}
%%%%%%%%%%%%%%%%%%%%%%%%%%%%% END Sample Tweets %%%%%%%%%%%%%%%%%%

Influencer marketing is dominated by \textit{native advertising} where there is no obvious distinction between \emph{commercial} (i.e., content that is monetized) and \emph{non-commercial} content such as personal thoughts, sentiment and experiences \cite{chia2012welcome}. Even though the disclosure of \emph{commercial} content (via keywords such as \textit{\#ad}, \textit{\#sponsored}) by influencers has become a requirement in some countries due to consumer protection obligations,\footnote{\url{https://icas.global/advertising-self-regulation/influencer-guidelines/}} identifying \emph{commercial} content in influencer posts is challenging in practice because (1) disclosure guidelines are not always followed, e.g., not including or hiding standard disclosure terms\footnote{Only about 10\% of affiliate marketing content on Pinterest and YouTube contains any disclosures \cite{mathur2018endorsements}.}\cite{wojdynski2016native,boerman2016informing,mathur2018endorsements,alassani2019product,de2020influencer}; and (2) brand cues (i.e., elements that may affect buying behavior) may appear in different modalities such as text, images or both \citep{sanchez-villegas-etal-2021-analyzing}. Figure \ref{fig:sampleLabels} shows an example of a \emph{commercial} and a \emph{non-commercial} post. Both examples appear to include products, however only the top example is \emph{commercial}. This makes it difficult for the users to distinguish between paid promotion and personal opinions. 

Therefore, automatically detecting whether an influencer's post involves paid promotion of products or services is of utmost importance for addressing issues related to transparency and regulatory compliance, such as misleading advertising or undisclosed sponsorships in large scale \citep{mathur2018endorsements,evans2017disclosing,wojdynski2018measuring,ducato2020one,ershov2020effects}. 
Previous work on identifying influencer commercial content has focused on analyzing user features (e.g., popularity and engagement) and network characteristics of influencers \citep{zarei2020characterising,kim2021discovering}, while the use of language and its relationship to images has not been explicitly explored. 

In this work, we present a new expert annotated Twitter (now X) dataset and an extensive empirical study on influencer multimodal content focused on analyzing the contribution of text and image modalities to \emph{commercial} and \emph{non-commercial} posts. Our main contributions are as follows: 
 \begin{itemize}
    \item We present a large publicly available dataset of $14,384$ text-image pairs and $1,614$ text-only influencer tweets written in English. Tweets are mapped into \emph{commercial} and \emph{non-commercial} categories; %via two ways: (1) automatic keyword-based labeling which is suitable in a real-world scenario with limited resources; (2) we include $3,049$ tweets annotated by experts in online advertising regulatory domain to ensure a high quality dataset for evaluation;
     \item We benchmark an extensive set of state-of-the-art language, vision and multimodal models for automatically identifying \emph{commercial} content, including prompting large language models (LLMs); 
     \item We propose a simple yet effective cross-attention multimodal approach that outperforms all text, vision and multimodal models;
     \item We conduct a qualitative analysis to shed light on the limitations of automatically detecting \emph{commercial} content, and provide insights into when each modality is beneficial.
\end{itemize}

%%%%%%%%%%%%%%%%%%%%%%%%%%%%%%%%%%%%%%%%%%%%%%%%%%%%%%%%%%%%%%

%%%%%%%%%%%%%%%%%%%%%%%%%%%%%%%%%%%%%%%%%%%%%%%%%%%%%%%%%%%%%%%%%%%%%%%%%%%%

% \usepackage[normalem]{ulem}

\begin{table*}[t!]
\centering
\small
\resizebox{\textwidth}{!}{
\begin{tabular}{|l|c|c|c|c|c|c|l|l|l|}
\hline
\textbf{Dataset}                & \textbf{\begin{tabular}[c]{@{}c@{}}Publicly \\ Available\end{tabular}} & \textbf{\begin{tabular}[c]{@{}c@{}}Posts w/o\\ brand\\ mentions\end{tabular}} & \textbf{\begin{tabular}[c]{@{}c@{}}Human\\ Annotation\end{tabular}} & \textbf{\begin{tabular}[c]{@{}c@{}}Keyword\\ Matching\end{tabular}} & \textbf{\begin{tabular}[c]{@{}c@{}}No. of \\ Commercial\\ Keywords\end{tabular}} & \textbf{Platform} & \multicolumn{1}{c|}{\textbf{Modality}} & \multicolumn{1}{c|}{\textbf{\begin{tabular}[c]{@{}c@{}}Time\\ Range\end{tabular}}} & \multicolumn{1}{c|}{\textbf{Domains}}                                                                       \\ \hline
\citet{10.1145/3449227}         & \xmark                                                                 & \xmark                                                                        & \xmark                                                              & \xmark                                                              & 0                                                                             & Twitter           & Text                                   & not specified                                                                      & fashion                                                                                                     \\ \hline
\citet{zarei2020characterising} & \xmark                                                                 & \cmark                                                                        & \xmark                                                              & \cmark                                                              & 7                                                                             & Instagram         & Text                                   & \begin{tabular}[c]{@{}l@{}}Jul 2019  -\\ Aug 2019\end{tabular}                     & not specified                                                                                               \\ \hline
\citet{yang2019influencers}     & \xmark                                                                 & \xmark                                                                        & \xmark                                                              & \cmark                                                              & 3                                                                             & Instagram         & Text \& Image                          & not specified                                                                      & not specified                                                                                               \\ \hline
\citet{kim2021discovering}      & \cmark                                                                 & \cmark                                                                        & \xmark                                                              & \cmark                                                              & 3                                                                             & Instagram         & Text \& Image                          & not specified                                                                      & not specified                                                                                               \\ \hline
\citet{kim2020multimodal}       & \cmark                                                                 & \xmark                                                                        & \xmark                                                              & \cmark                                                              & 1                                                                             & Instagram         & Text \& Image                          & \begin{tabular}[c]{@{}l@{}}Oct 2018 -\\ Jan 2019\end{tabular}                      & \begin{tabular}[c]{@{}l@{}}beauty, family, food,\\ fashion, pet, fitness, \\ interior, travel,\end{tabular} \\ \hline
MICD (Ours)                     & \cmark                                                                 & \cmark                                                                        & \cmark                                                              & \cmark                                                              & 26                                                                            & Twitter           & Text \& Image                          & \begin{tabular}[c]{@{}l@{}}Jan  2015 - \\ Aug 2021\end{tabular}                    & \begin{tabular}[c]{@{}l@{}}beauty, travel, food\\ fitness, technology, \\ lifestyle\end{tabular}            \\ \hline
\end{tabular}
}%
\caption{A comparison of existing datasets for influencer content analysis}
\label{tab:dataset-comp}
\end{table*}
%%%%%%%%%%%%%%%%%%%%%%%%%%%%%%%%%%%%%%%%%%%%%%%%%%%%%%%%%%%%%%%%%%%%%%%%%%%%%

\section{Related Work}
\label{sec:rel-work}
\subsection{Computational Studies on Influencers} 

Previous work has analyzed the characteristics of influencers on social media platforms such as Twitter \cite{huang2014cross,lagree2018algorithms,10.1145/3449227}, Instagram \cite{kim2017social,kim2021evaluating,10059732} and Pinterest \citep{gilbert2013need,mathur2018endorsements}. \citet{kim2017social} investigate the social relationships and interactions among influencers while \citet{kim2021evaluating} explore the audience loyalty and content authenticity. On Twitter, \citet{lagree2018algorithms} leverage social network analysis to discover influencers that achieve high reach on advertising campaigns and \citet{10.1145/3449227} study the relationships among fashion influencers to understand who they follow, mention, and retweet. Using posts from Pinterest and YouTube, \citet{mathur2018endorsements} examine whether influencers comply with advertising disclosure regulations and show that while influencer commercial content has increased over the years, its disclosure remains limited. 

\subsection{Data Resources for Influencer Content Analysis}

Datasets for analyzing influencer content have been developed to analyze the influencers' impact on spreading information \citep{10.1145/3449227}, categorizing influencers into different domains, e.g., fashion, beauty \citep{kim2020multimodal},
and analyzing the characteristics of branded content \cite{yang2019influencers}. \citet{yang2019influencers} introduce a dataset to study how influencers mention brands in their posts. They collect 800K Instagram posts from $18$K influencers that explicitly mention (@mention) a brand, and characterize them as sponsored or non-sponsored using three sponsorship indicators: \textit{\#ad}, \textit{\#sponsored}, \textit{\#paidAD}. 

Datasets for analyzing commercial content shared by influencers have been developed by \citet{zarei2020characterising} and \citet{kim2021discovering}. \citet{zarei2020characterising} present a dataset consisting of 35K Instagram posts and 99K stories (i.e., posts that disappear after $24$ hours) from $12$K influencers and use an LSTM model \citep{hochreiter1997long} to identify whether a post is sponsored or not. \citet{kim2021discovering} develop a dataset of 38K influencer posts that explicitly mention (@mention) a brand. Similar to \citet{yang2019influencers}, they label these posts as sponsored if they contain at least one of three sponsorship indicators: \textit{\#ad}, \textit{\#sponsored}, \textit{\#paidAD}. They propose an attention-based neural network model to classify posts as sponsored or non-sponsored.

\paragraph{Limitations of existing resources} Table \ref{tab:dataset-comp} compares existing datasets for analyzing influencer content. We observe that current datasets have only used a limited set of keywords (e.g., \textit{\#ad}) for identifying posts with commercial content (seven or less). While some datasets include only text content \cite{zarei2020characterising}, others focus only on posts that explicitly mention (@mention) a brand \cite{yang2019influencers,kim2021discovering}. In contrast to prior datasets for analyzing influencer commercial content that use Instagram, we use Twitter  because it is a text-first platform and has rapidly increased in popularity as a tool for influencer marketing. For instance, $49$\% of Twitter users say that they have made a purchase as a direct result of a Tweet from an influencer.\footnote{\url{https://blog.twitter.com/en_us/a/2016/new-research-the-value-of-influencers-on-twitter}}

%%%%%%%%%%%%%%%%%%%%%%%%%%%%%%%%%%%%%%%%%%%%%%%%%%%%%%%%%%%%%%%%%%%%%%%%%%%%%
\section{Multimodal Influencer Content Dataset (MICD)}

We present a new multimodal influencer content dataset (MICD) consisting of Twitter posts mapped into \emph{commercial} and \emph{non-commercial} classes.

\subsection{Retrieving Candidate Influencers} 
To map tweets into these two classes, we first need to identify candidate influencers on Twitter. We look for candidate accounts in six different domains (i.e., \emph{Beauty}, \emph{Travel}, \emph{Fitness}, \emph{Food}, \emph{Tech} and \emph{Lifestyle}) to ensure thematic diversity. The domains related to `Beauty', `Fitness', `Travel' and `Lifestyle' are among the most popular in Twitter,\footnote{\url{https://influencermarketinghub.com/influencer-marketing-benchmark-report-2021/}} while \emph{Food} and \emph{Tech} have recently gained attention \citep{alassani2019product,weber2021s}. To retrieve influencers, we query for accounts that contain domain-specific keywords in their bios (e.g., \textit{beauty vlogger}, \textit{travel influencer}, \textit{lifestyle blogger}, \textit{food writer}) as influencers tend to provide such information in profile descriptions \citep{kim2020multimodal}.\footnote{Influencer accounts were manually validated to ensure bots are not included.} We collect all available image-text tweets written in English from each account using the Academic Twitter API.\footnote{\url{https://developer.twitter.com/en/products/twitter-api}} Duplicate tweets with identical text are removed. 

\subsection{Keyword-based Weak Labeling} 
\label{sec:datalbls}

We initially use a keyword-based strategy to automatically map posts into the \emph{commercial} and \emph{non-commercial} categories (i.e., weak labeling). This is suitable in a real-world scenario of an automatic regulatory compliance system with limited resources for manually labeling all available posts \citep{zarei2020characterising,kim2021discovering}. 

\paragraph{Commercial}
Commercial tweets include content that promotes or endorses a brand or its products or services, a free product or service or any other incentive. Thus, we extract keywords strongly associated with influencer marketing following the official guidelines provided by the Federal Trade Commission \citep{ftc} in the US, and the Advertisements Standards Authority and Competition and Markets Authority in the UK \citep{cma}. These guidelines contain lists of keywords to appropriately disclose commercial content. In this work, we considerably extend the keyword lists (extended and verified by members of a national consumer authority) to not only include recommended sponsorship disclosure terms (e.g., \textit{\#ad}, \textit{\#sponsored}), but also terms that are relevant to different business models (i.e., market practices based on the obligations of the parties) such as gifting (e.g., \textit{\#gift}, \textit{\#giveaway}), endorsements (e.g., \textit{\#ambassador}) and affiliate marketing (e.g., \textit{\#aff}, \textit{discount code}). A complete list of keywords can be found in Appx. \ref{sec:appendix_keywords}. We label as \emph{commercial} all tweets containing at least one of the influencer marketing keywords excluding tweets where the keyword is negated (e.g., \textit{not ad}, \textit{not an ad}). To avoid data leakage in the experiments, we remove all of the keywords used for data labeling (see Sec. \ref{sec:txtproc}) from the posts after labeling them. As a result, our models can identify \emph{commercial} content without the use of such terms (see Sec. \ref{sec:method}).

\paragraph{Non-commercial}
Non-commercial posts refer to organic content such as personal ideas, comments and life updates that do not aim for monetization. Thus, all tweets that do not include any of the keywords presented above are considered \emph{non-commercial}. To balance the dataset, we sample \emph{non-commercial} posts weighted according to the number of \emph{commercial} tweets for each account.

\subsection{Data Splits} 
\label{sec:data-splits}

\paragraph{Text-Image Sets} We split the tweets into train, dev and test sets at the account level (i.e., tweets included in each split belong to different accounts) to ensure that models can generalize to unseen influencer accounts and prevent information leakage in our experiments. 

\paragraph{Text-only Test Set} We further collect text-only posts from influencer accounts in the test set. We sample text-only tweets according to the number of tweets for each influencer account in the test set, resulting in a total of $1,614$ text-only tweets. This is done to account for cases where only text content is provided.\footnote{Note that while text-only tweets are prevalent on Twitter, image-only tweets are uncommon.}
\begin{table}[t!]
\centering
\small
\begin{tabular}{|l|c|c|c|c|}
\hline
%\rowcolor[HTML]{C0C0C0} 
\textbf{Domain} & \textbf{Accounts}% & \textbf{Common Objects in Images}
\\\hline
Beauty           & 22  
\\ \hline
Travel           & 22   

\\ \hline
Fitness         & 15   

\\ \hline
Food            & 22   

\\ \hline
Tech            & 20    

\\ \hline
Lifestyle        & 31   

\\ \hline
Total            &132        

\\ \hline
\end{tabular}
\caption{Number of influencer accounts by domain}
\label{tab:spTweets0}
\end{table}
\subsection{Human Data Annotation}
To ensure a high quality data set for evaluation, we use human annotators for labeling all tweets in both test sets (text-image and text-only test sets).\footnote{We received approval from the Ethics Committee of our institution. Annotation guidelines can be found in Appx. \ref{appdx:annotation}.} Four volunteer annotators from our institution, each with a substantial legal background and knowledge of advertising disclosure regulations labeled the test dataset. A workshop was held to introduce the task to the annotators, explain the annotation guidelines and run a calibration round on a random set of $20$ examples. All tweets in the test sets were labeled by two different annotators as \emph{commercial}, \emph{non-commercial}, or \emph{unclear} (i.e., it is not clear whether the post contains \emph{commercial} content or not). In cases of disagreement, a third independent annotator assigned the final label (\emph{commercial} or \emph{non-commercial}) after adjudication. Posts labeled as \emph{unclear} ($15$) are removed, as well as posts written in other language than English ($2$). 

The inter-annotator agreement between two annotations across all tweets is $0.78$ Cohen's-Kappa \citep{cohen1960coefficient} that corresponds to the upper part of the \textit{substantial} agreement band \citep{artstein2008inter}. Furthermore, the agreement between the automatic weak labels and the resulting human annotations is $0.67$ Cohen's-Kappa which corresponds to \textit{substantial} agreement and denotes weak labels of good quality for model training.  

Our final dataset contains $14,384$ text-image pairs ($7,259$ \emph{non-commercial} and $7,125$ \emph{commercial}). Additionally, the text-only test set consists of $1,614$ tweets ($1,377$ \emph{non-commercial} and $237$ \emph{commercial}). Table \ref{tab:splits} shows the distribution of \emph{commercial} and \emph{non-commercial} tweets by split.

\begin{table}[t!]
\centering
\small
\resizebox{0.48\textwidth}{!}{%
\begin{tabular}{|c|c|c|c|}
\hline
%\rowcolor[HTML]{C0C0C0} 
 \textbf{Split} & \textbf{Non-commercial} & \textbf{Commercial} &  \textbf{Total} \\ \hline

%\cellcolor[HTML]{EFEFEF}
\textbf{Train} &        5,781     & 5,596   & 11,377 (79.1\%)           
\\ \hline
%\cellcolor[HTML]{EFEFEF}
\textbf{Dev}   &        789       & 783     & 1,572 (10.9\%)           
\\ \hline
%\cellcolor[HTML]{EFEFEF}
\textbf{Test} &         689       & 746     & 1,435 (10\%)  
\\ \hline
\textbf{Total}  & 7,259	   & 7,125   & 14,384           
\\ \hline \hline
\textbf{Text-only Test} &    1,377      & 237     & 1,614  
\\ \hline \hline
\textbf{All} &    8,636      & 7,352     & 15,998  
\\ \hline
%\cellcolor[HTML]{EFEFEF}
\end{tabular}
}
\caption{Dataset statistics showing the number of tweets for each split.}
\label{tab:splits}
\end{table}

\subsection{Exploratory Analysis}
Exploratory analysis of our dataset revealed that influencer accounts in our dataset have between 8K and 500K followers covering micro and macro influencers which are considered to create highly persuasive content \citep{kay2020less}. Table \ref{tab:spTweets0} shows the number of influencer accounts per domain. In average, each domain contains $22$ accounts, and all accounts have a minimum of 10 \emph{commercial} tweets. Finally, we observe a different label distribution in text-image and text-only test splits. Text-only test split is unbalanced with most posts manually annotated as \emph{non-commercial} (85.32\% \emph{non-commercial}, 14.68\% \emph{commercial}). On the other hand, text-image test set label distribution is balanced (48.01\% \emph{non-commercial}, 51.99\% \emph{commercial}). This highlights the use of visuals in influencer marketing for effectively advertising products, which is consistent with findings in conventional online advertising research \citep{mazloom2016multimodal}. It also emphasizes the multimodal nature of the task. 

\subsection{Comparison with Related Datasets}

Table \ref{tab:dataset-comp} compares our dataset, MICD, to related datasets for influencer content analysis (see Sec. \ref{sec:rel-work}). Our dataset contains posts with and without explicit (i.e., @USER) brand mentions from influencers of different domains. We follow a similar approach for weak labeling \emph{commercial} posts as previous work \cite{zarei2020characterising,kim2021discovering}, but we considerably extend the list of keywords following relevant guidelines and experts feedback (see Sec. \ref{sec:datalbls}). Moreover, we include test sets with a total of $3,049$ tweets annotated by experts in the legal domain. We anticipate that this dataset will be beneficial not only for this study, but also for future influencer content analysis research. 

%%%%%%%%%%%%%%%%%%%%%%%%%%%%%%%%%%%%%%%%%%%%%%%%%%%%%%%%%%%%%%%%%%%%%

\section{Influencer Content Classification Models}
\label{sec:method} 

Given a social media post $P$ (e.g., a tweet) consisting of a text and image pair ${(L,I)}$, the task is to classify a post $P$ into the correct category (\emph{commercial} or \emph{non-commercial}). %For the purpose of our study, we focus on influencer language and its interplay with images. 

\subsection{Unimodal Models}

\paragraph{Prompting} We first experiment with prompting \textbf{Flan-T5} \citep{chung2022scaling} and \textbf{GPT-3} \cite{brown2020language}. We use the following prompt: \textit{``Label the next text as `commercial' or `not commercial'. Text: <TWEET>"}. We map responses to the corresponding \emph{commercial} or \emph{non-commercial} class and report results for each model (zero-shot). We further experiment with few-shot prompting by appending four randomly selected training examples\footnote{Appx. \ref{appdx:prompting} includes the template we use for these prompts.} (two examples from each class) before each prompt (few-shot). We run this three times with a different set of examples and report average performance.

\paragraph{Image-only Models} We fine-tune two pre-trained models that achieve state-of-the-art results in various computer vision classification tasks by adding an output classification layer: (1) \textbf{ResNet}152 \cite{he2016deep} and (2) \textbf{ViT} \cite{dosovitskiy2020image}. ResNet uses convolution to
aggregate information across locations, while ViT uses self-attention for this purpose. Both models are pre-trained on the ImageNet dataset \cite{russakovsky2015imagenet}.

\paragraph{Text-only Recurrent Model} \citet{zarei2020characterising} propose a contextual Long-Short Term Memory (LSTM) neural network architecture for identifying posts in Instagram. Thus, we also experiment with a similar bidirectional LSTM network with a self-attention mechanism \citep{hochreiter1997long} to obtain the tweet representation that is subsequently passed to the output layer with a softmax activation function (\textbf{BiLSTM-Att}).

\paragraph{Text-only Transformers} We fine-tune %\footnote{We add a classification layer on top of the [CLS] token.} 
two pre-trained transformer-based~\cite{Vaswani2017} models for commercial posts prediction: \textbf{BERT} \cite{devlin-etal-2019-bert} and \textbf{BERTweet} \citep{nguyen-etal-2020-bertweet} by adding a classification layer on top of the [CLS] token. \textbf{BERTweet} is a BERT based model pre-trained on a large-scale corpus of English Tweets.

\subsection{Multimodal Models}

\paragraph{Text \& Image Transformers} We fine-tune three multimodal transformer-based models: \textbf{MMBT}~\citep{kiela2019supervised}, \textbf{ViLT}~\citep{kim2021vilt} and \textbf{LXMERT}~\citep{tan-bansal-2019-lxmert}. MMBT uses ResNet and BERT as image and text encoders respectively, ViLT uses a convolution-free encoder similar to ViT, and LXMERT takes \textit{object-level} features as input (see Sec. \ref{sec:txtproc}). ViLT and LXMERT are multimodally pre-trained on visual-language tasks such as image-text matching and and visual question answering.

\paragraph{Aspect-Attention} \citet{kim2021discovering} proposed an aspect-attention fusion model to rank Instagram posts based on their likelihood of including undeclared paid partnerships. Thus, we repurpose their model to identify commercial posts on Twitter. Aspect-attention fusion consists of  generating a score for each modality by applying the attention mechanism across the image and text vectors. Then, the multimodal post representation is produced by computing a linear combination of the score and the unimodal representations. The model is fine-tuned by adding a fully-connected layer with a softmax activation function (\textbf{Aspect-Att}).

\paragraph{ViT-BERTweet-Att} We propose to combine unimodal pretrained representations via cross-attention fusion strategy so that text features can guide the model to pay attention to the relevant image regions. We use BERTweet to obtain contextual representations of the text content $L \in R^{d_L \times m_L}$, where $L$ is the output of the last layer of BERTweet, $d_L$ is the hidden size of BERTweet and $m_L$ is the text sequence length. For encoding the images, we use the Vision Transformer pre-trained on ImageNet \citep{russakovsky2015imagenet}. We obtain the visual representations of the image content $I \in R^{d_I \times m_I}$, where $I$ is the output of the last layer of ViT, $d_I$ is the hidden size of ViT and $m_I$ is the image sequence length. We propose to capture the inter-modality interactions using a cross-attention layer. Specifically, given $L$ and $I$, we compute the scaled dot attention with $L$ as queries, and $I$ as keys and values as follows: \small $\text{Cross-Att}(L,I) = \text{softmax}(\frac{[W^Q L] [W^K I]^T}{\sqrt{d_k}}) [W^V I]$,
% \tiny
% \begin{equation*}
%     Att(L,I) = softmax(\frac{[W^Q L] [W^K V]^T}{\sqrt{d_k}}) [W^V I]
% \end{equation*}
\normalsize
where \small $\{W^Q,W^K,W^V\}$ \normalsize are learnable parameters, \small $d_k=d^L=d^I$\normalsize, and \small$\text{Cross-Att}(L, I)$ $\in R^{m_L \times d_k}$\normalsize 

The multimodal representation vector $h$ is obtained by concatenating the `classification' [CLS]$_L$ token from $L$ (output from the last layer of BERTweet), and the [CLS]$_{Att}$ token from the output of the cross-attention layer (\small$\text{Cross-Att}(L,I)$\normalsize). In this way, we leverage the text content of the influencer posts, and the relevant information from the image content. We fine-tune the model on the commercial content classification task by adding a fully-connected layer with a softmax activation function.\footnote{Figure \ref{fig:model} shows a diagram of the model.}

%%%%%%%%%%%%%%%%%%%%%%%%%%%%%%%%%%%%%%%%%%%%%%%
\section{Experimental Setup}
\label{sec:expsetup}

\subsection{Data Processing}
\label{sec:txtproc}
\paragraph{Text} For each tweet, we lowercase and tokenize text using DLATK \cite{schwartz-etal-2017-dlatk}. We also replace URLs and user @-mentions with placeholder tokens following the BERTweet pipeline \citep{nguyen-etal-2020-bertweet}. Emojis are replaced with their corresponding text string, e.g thumbs\_up. Keywords used in the weak labeling process (Sec. \ref{sec:datalbls}) are removed from all \emph{commercial} tweets.

\paragraph{Image} Images are resized to ($224\times224$) pixels representing a value for the red, green and blue color in $[0,255]$. The pixel values are normalized to $[0-1]$. For LXMERT, we extract \textit{object-level} features using Faster-RCNN \cite{ren2016faster} as in \citet{anderson2018bottom} and keep $36$ objects for each image as in \citet{tan-bansal-2019-lxmert}. 

\subsection{Most Freq. Baseline and Evaluation}

\paragraph{Most Freq. Baseline} We assign the most frequent label in the training set to all instances in the test set.

\paragraph{Evaluation} We evaluate all models using weighted-averaged\footnote{Macro-averaged results are included in Appx. \ref{sec:appendix_res}.} F1, precision, and recall to manage imbalanced classes. Results are obtained over three runs using different random seeds reporting average and standard deviation. %Details on hyperparameter selection are included in Appx. \ref{sec:appendix_hp}.

\subsection{Implementation Details}
\label{sec:appendix_hp}
We select the hyperparameters for all models using early stopping by monitoring the validation loss. We use the Adam optimizer~\citep{kingma2014adam}. We estimate the class weights using the `balanced' heuristic \cite{king2001logistic}. All experiments (unless indicated) are performed using an Nvidia V100 GPU with a batch size of 16. 

%\subsection{Implementation Details}
\paragraph{Prompting} We use one GPU T4 to obtain the inference results from Flan-T5 \citep{chung2022scaling} model. We use the large version from HuggingFace library (780M parameters) \citep{wolf2019huggingface}. For GPT-3 \cite{brown2020language}, we use the \textit{text-davinci-003} model via the OpenAI\footnote{\url{https://platform.openai.com/docs/}} Library. %(\$54 USD total). 
Prompt templates are included in Appx. \ref{appdx:prompting}.

%\subsection{Unimodal Models} 
\paragraph{Image-only} For ResNet152 \citep{he2016deep}, we fine-tune for 1 epoch with learning rate $\eta=1e^{-5}$ and dropout $\delta=0.05$ before passing the image representation through the classification layer. We fine-tune ViT \citep{dosovitskiy2020image} for 3 epochs with learning rate $\eta=1e^{-5}$ and dropout $\delta=0.05$. $\eta \in \{1e^{-3},1e^{-4},1e^{-5}\}$ and $\delta$ in $[0,0.5]$, random search.

\paragraph{Text-only Recurrent Model} For BiLSTM-Att we use 200-dimensional GloVe embeddings \cite{pennington-etal-2014-glove} pre-trained on Twitter data. The maximum sequence length is set to 50. The LSTM size is $h = 32$ where $h \in \{32,64,100\}$ with dropout $\delta = 0.3$ where $\delta \in [0,0.5]$, random search. We use Adam \cite{kingma2014adam} with learning rate $\eta=1e^{-3}$ with $\eta \in \{1e^{-3},1e^{-4},1e^{-5}\}$, minimizing the binary cross-entropy using a batch size of 8 over 6 epochs with early stopping.

\paragraph{Text-only Transformers} We fine-tune BERT and BERTweet for 20 epochs and choose the epoch with the lowest validation loss. We use the pre-trained base-uncased model for BERT \cite{Vaswani2017, devlin-etal-2019-bert} from HuggingFace library (12-layer, 768-dimensional) \citep{wolf2019huggingface}, and the base model for BERTweet \citep{nguyen-etal-2020-bertweet} with a maximal sequence length of 128. We fine-tune BERT for 1 epoch, learning rate $\eta=1e^{-5}$ and dropout $\delta=0.05$; and BERTweet for 2 epochs, $\eta=1e^{-5}$ and $\delta=0.05$. For all models $\eta \in \{2e^{-5},1e^{-4},1e^{-5}\}$ and $\delta \in [0,0.5]$, random search.

%\subsection{Multimodal Models}

\paragraph{Text \& Image Transformers} We train MMBT~\citep{kiela2019supervised} for 1 epoch and $\eta=1e^{-5}$ where $\eta \in \{1e^{-3},1e^{-4},1e^{-5}\}$ and dropout $\delta =0.05$ ($\delta$ in $[0,0.5]$, random search) before passing through the classification layer. ViLT~\citep{kim2021vilt} is fine-tuned for 4 epochs and $\eta=1e^{-5}$, vision layers are frozen. LXMERT~\citep{tan-bansal-2019-lxmert} is fine-tuned for 3 epochs with $\eta=1e^{-5}$ and $\delta=0.05$.

\paragraph{Aspect-Attention and ViT-BERTweet-Att} We train Aspect-Attention and ViT-BERTweet-Att with BERTweet as text encoder and ViT as image encoder for 15 epochs and choose the epoch with the lowest validation loss. Aspect-Attention: 1 epoch with $\eta=1e^{-5}$ and $\delta=0.05$ and ViT-BERTweet-Att 3 epochs with $\eta=1e^{-5}$ and $\delta=0.05$ ; The dimensionality of the multimodal representation is 768. $\eta \in \{1e^{-3},1e^{-4},1e^{-5}\}$ and $\delta$ in $[0,0.5]$, random search.

%%%%%%%%%%%%%%%%%%%%%%%%%%%%%%%%%%%%%%%%%%%%%%%%%%%%%%%%%%%%%%%%%%%%%%%%%%

\begin{table}[t!]
\centering
\small
\resizebox{\columnwidth}{!}{
\begin{tabular}{|l|l|c|c|} %
\hline
\multicolumn{1}{|c|}{
\textbf{Model}}&\multicolumn{1}{|c|}{
\textbf{F1}} & \multicolumn{1}{|c|}{\textbf{P}} & \multicolumn{1}{|c|}{\textbf{R}} 
\\ \hline
Most Freq.                 & 31.15$_{0.0}$ & 23.05$_{0.0}$ & 48.01$_{0.0}$  \\ \hline \hline
\textbf{Prompting} &&&\\
Flan-T5 (zero-shot)                   & 42.98$_{0.0}$ & 72.01$_{0.0}$ & 53.51$_{0.0}$ \\
Flan-T5 (few-shot)                    & 48.70$_{1.6}$ & 62.07$_{0.9}$ & 53.47$_{0.6}$ \\
GPT-3 (zero-shot)                     & 63.91$_{0.0}$ & 65.64$_{0.0}$ & 64.81$_{0.0}$\\
GPT-3 (few-shot)                      &69.57$_{1.5}$&71.69$_{2.1}$&70.01$_{0.8}$\\
\hline \hline
\textbf{Image-only} &&&\\
ResNet           & 59.59$_{0.5}$ &   59.85$_{0.5}$   & 59.60$_{0.5}$  \\ 
ViT              & 60.81$_{1.3}$ &   61.58$_{0.9}$  & 61.02$_{1.2}$   \\ 
\hline \hline
\textbf{Text-only} &&&\\
BiLSTM-Att$^*$ \cite{zarei2020characterising}          & 66.10$_{0.7}$ & 66.48$_{0.8}$ & 65.15$_{0.7}$  \\ 
BERT             & 74.32$_{0.6}$  &  75.01$_{0.6}$   &  74.43$_{0.7}$ \\ 
BERTweet         & 76.34$_{0.3}$  &  76.80$_{0.3}$   & 76.45$_{0.3}$  \\
\hline \hline
\textbf{Text \& Image} &&&\\
ViLT            & 68.46$_{0.9}$  &  66.66$_{3.8}$    & 66.66$_{3.8}$  \\
LXMERT          & 70.64$_{0.4}$  &  71.00$_{0.3}$    & 70.68$_{0.4}$ \\ 
MMBT            & 73.58$_{0.4}$  &  73.79$_{0.6}$    & 73.59$_{0.4}$  \\
Aspect-Att$^*$ \cite{kim2021discovering}                & 75.45$_{0.8}$ &  77.42$_{1.1}$ & 75.68$_{0.7}$  \\ 
ViT-BERTweet-Att (Ours)   & {\bf 77.50}$^\ddagger_{0.6}$   &  {\bf78.46}$^\dagger_{0.5}$    &  {\bf77.61}$^\ddagger_{0.6}$ \\
\hline
\end{tabular}
}
\caption{Weighted F1-Score, precision (P) and recall (R) for commercial influencer content prediction. $\dagger$ and $\ddagger$ indicates statistically significant improvement (t-test, $p<0.05$) over BERTweet, and both BERTweet and Aspect-Att respectively. %We include standard deviations in Appx. \ref{sec:appendix_res}. 
$^*$ denotes current state-of-the-art models for influencer commercial content detection. Subscripts denote standard deviations. Best results are in bold.}
\label{tab:results}
\end{table}

\section{Results}
\label{sec:res}
Table \ref{tab:results} presents the performance on \emph{commercial} and \emph{non-commercial} influencer content prediction of all predictive models on our new multimodal influencer content dataset (MICD).

\subsection{Unimodal Models}

We first observe that the two image-only models obtain similar performance. Although both models surpass Most Freq. baseline and Flan-T5 prompting, the text-only models (BiLSTM-ATT, BERT and BERTweet) perform better than image-only models. This corroborates results from previous work in multimodal computational social science \citep{wang-etal-2020-cross-media,ma-etal-2021-effectiveness} and influencer content analysis \citep{kim2021discovering}. 
We further note that BERT-based models (BERT and BERTweet) outperform GPT-3 prompting and  BiLSTM-Att models over $4\%$ across all metrics. Among the text-only models, BERTweet achieves the highest performance with $76.34$, $76.80$ and $76.45$ weighted F1, precision and recall respectively. 

\begin{table}[t!]
\centering
\small
\resizebox{\columnwidth}{!}{
\begin{tabular}{|l|c|c|c|} 
\hline
\multicolumn{1}{|c|}{
\textbf{Model}}&\multicolumn{1}{|c|}{
\textbf{F1}} & \multicolumn{1}{|c|}{\textbf{P}} & \multicolumn{1}{|c|}{\textbf{R}} 
\\ \hline
BERTweet                 & 76.34$_{0.3}$  &  76.80$_{0.3}$  & 76.45$_{0.3}$  \\ 
ViT                      & 60.81$_{1.3}$ &   61.58$_{0.9}$  & 61.02$_{1.2}$  \\
ViT-BERTweet-Concat      & 76.34$_{0.9}$  &  78.10$_{0.5}$  & 76.54$_{0.8}$   \\
ViT-BERTweet-Att (Ours)  &\textbf{ 77.50}$_{0.6}$ & \textbf{ 78.46}$_{0.5}$ & \textbf{77.61}$_{0.6}$ \\
\hline
\end{tabular}
}
\caption{Comparison of each of the ViT-BERTweet-Att components including the removal of the Cross-Att layer (ViT-BERTweet-Concat). Subscripts denote standard deviations. Best results are in bold.}
\label{tab:res_ablation}
\end{table}

\subsection{Multimodal models} 

State-of-the-art pre-trained multimodal models, ViLT and LXMERT fail to outperform text-only transformers achieving only $68.46$ and $70.64$ weighted F1 respectively. This emphasizes the challenges for modeling multimodal influencer content. Specifically, ViLT and LXMERT are pretrained on standard vision-language tasks including image captioning and visual question answering \citep{Zhou_Palangi_Zhang_Hu_Corso_Gao_2020, NEURIPS2019_c74d97b0} using data where text and image modalities share common semantic relationships. In contrast, social media advertising frequently employs various types of visual and text rhetoric (e.g., symbolism) to convey their message with no obvious relationship between text and image \citep{vempala-preotiuc-pietro-2019-categorizing,hessel-lee-2020-multimodal,sanchez-villegas-aletras-2021-point}. Similar behavior is observed with MMBT which obtains comparable performance to BERT. This suggests it is more beneficial to use a text-only encoder (BERTweet) that has been pre-trained on the same domain, in this case Twitter, than fine-tuning a more complex out-of-the-box multimodal transformer model (e.g., ViLT, LXMERT, MMBT). 

BERTweet and ViT are used by Aspect-Att (a state-of-the-art model for influencer commercial content prediction) and our model, ViT-BERTweet-Att, to obtain text and visual representations. However, only ViT-BERTweet-Att outperforms all text- and image-only models ($77.50$, $78.46$, $77.61$ weighted F1, precision, and recall), indicating that not only the choice of text and image encoders is important, but so is the fusion strategy for effectively modeling text-image relationships for identifying influencer \emph{commercial} content.

%%%%%%%%%%%%%%%%%%%%%%%%%%%%%%%%%%%%%%%%%%%%%%%%%%%%%%%%%%%%%%%%%%%%%%%%%
\subsection{Ablation Study}

To analyze the contribution of each component of our ViT-BERTweet-Att in identifying \emph{commercial} posts, Table \ref{tab:res_ablation} shows the performance of ViT, BERTweet, and ViT-BERTweet-Att with and without the Cross-Att layer (see Sec. \ref{sec:method}). ViT-BERTweet-Att without the Cross-Att layer consists of simply concatenating text and image vectors (\textbf{ViT-BERTweet-Concat}). While the performance of BERTweet and ViT-BERTweet-Concat are comparable (BERTweet and ViT-BERTweet-Concat weighted F1: $76.34$), ViT-BERTweet-Att (weighted F1: $77.50$) outperforms BERTweet suggesting the Cross-Att layer successfully captures the relevant regions in images for identifying \emph{commercial} posts.

%%%%%%%%%%%%%%%%%%%%%%%%%%%%%%%%%%%%%%%%%%%%%%%%%%%%%%%%%%%%%%%%%%%%%%%%%
\subsection{Text-only Test Set Evaluation}
Finally, previous work on text-image classification in \emph{commercial} influencer content has only experimented with fully paired data where every post contains an image and text \cite{kim2021discovering}. However, this requirement may not always hold since not all posts contain both modalities. Thus, we further evaluate our models on our text-only test set (see Sec. \ref{sec:data-splits}). Table \ref{tab:res_txt_only} shows the results obtained. We observe a consistent improvement of ViT-BERTweet-Att multimodal model over BERTweet text-only model, i.e., $88.69$ versus $87.50$. This suggests that multimodal modeling of influencer posts is beneficial for identifying text-only \emph{commercial} posts.

\begin{table}[t!]
\centering
\small
\resizebox{\columnwidth}{!}{
\begin{tabular}{|l|c|c|c|} %
\hline
\multicolumn{1}{|c|}{
\textbf{Model}}&\multicolumn{1}{|c|}{
\textbf{F1}} & \multicolumn{1}{|c|}{\textbf{P}} & \multicolumn{1}{|c|}{\textbf{R}} 
\\ \hline
Most Freq.                 & 78.55$_{0.0}$ & 72.78$_{0.0}$  & 85.31$_{0.0}$   \\ 
Flan-T5 (zero-shot)        & 81.02$_{0.0}$ & 80.41$_{0.0}$  & 84.88$_{0.0}$   \\
Flan-T5 (few-shot)         & 82.22$_{0.5}$ & 81.72$_{0.6}$ & 83.56$_{0.6}$    \\
GPT-3   (zero-shot)        & 77.26$_{0.0}$ & 85.12$_{0.0}$ & 73.79$_{0.0}$    \\
GPT-3   (few-shot)         & 84.03$_{3.0}$ & 85.55$_{1.1}$  & 83.68$_{4.8}$   \\
BERTweet                   & 87.50$_{1.0}$ & 88.58$_{0.4}$  & 86.84$_{1.3}$   \\
ViT-BERTweet-Att (Ours)    & \textbf{88.69}$_{0.2}$ & \textbf{88.69}$_{0.2}$  & \textbf{88.93}$_{0.5}$   \\
\hline
\end{tabular}
}
\caption{Weighted F1-Score, precision (P) and recall (R) for commercial influencer content prediction for tweets containing text only. %We include standard deviations in Appx. \ref{sec:appendix_res}.
Subscripts denote standard deviations. Best results are in bold.}
\label{tab:res_txt_only}
\end{table}

%%%%%%%%%%%%%%%%%%%%%%%%%%%%%%%%%%%%%%%%%%%%%%%%%%%%%%%%%%%%%%%%%%%%%

\section{Qualitative Analysis}
\label{sec:analysis}

We finally perform a qualitative analysis of the classification effectiveness between ViT-BERTweet-Att and the best text-only model (BERTweet). We analyze the strengths and limitations of each model. % when each modality contributes to the task and uncover their main limitations. 

\paragraph{Multimodal modeling helps to reduce the number of false positives.} We find that $53\%$ of BERTweet errors from the text-image test set are false positives, i.e., misclassifying \emph{non-commercial} posts as \emph{commercial}, which would be problematic for an automated regulatory compliance system. Our multimodal model, ViT-BERTweet-Att, on the other hand, correctly classifies $38\%$ of BERTweet's false positive mistakes such as the \emph{non-commercial} post in Figure \ref{fig:sampleLabels}. Similarly, for text-only posts, we observe that $69\%$ of BERTweet missclassifications correspond to false positive errors. $50.9\%$ of these posts are correctly classified by ViT-BERTweet-Att. 

\begin{figure}[!t]
    %\footnotesize
    \centering
    %\small
     \tiny
    \begin{tabular}{|m{2.15cm}|m{1.99cm}|m{2.25cm}|}
    \textbf{Post A} & \textbf{Post B} & \textbf{Post C} \\
    \hline
     \includegraphics[scale=0.03]{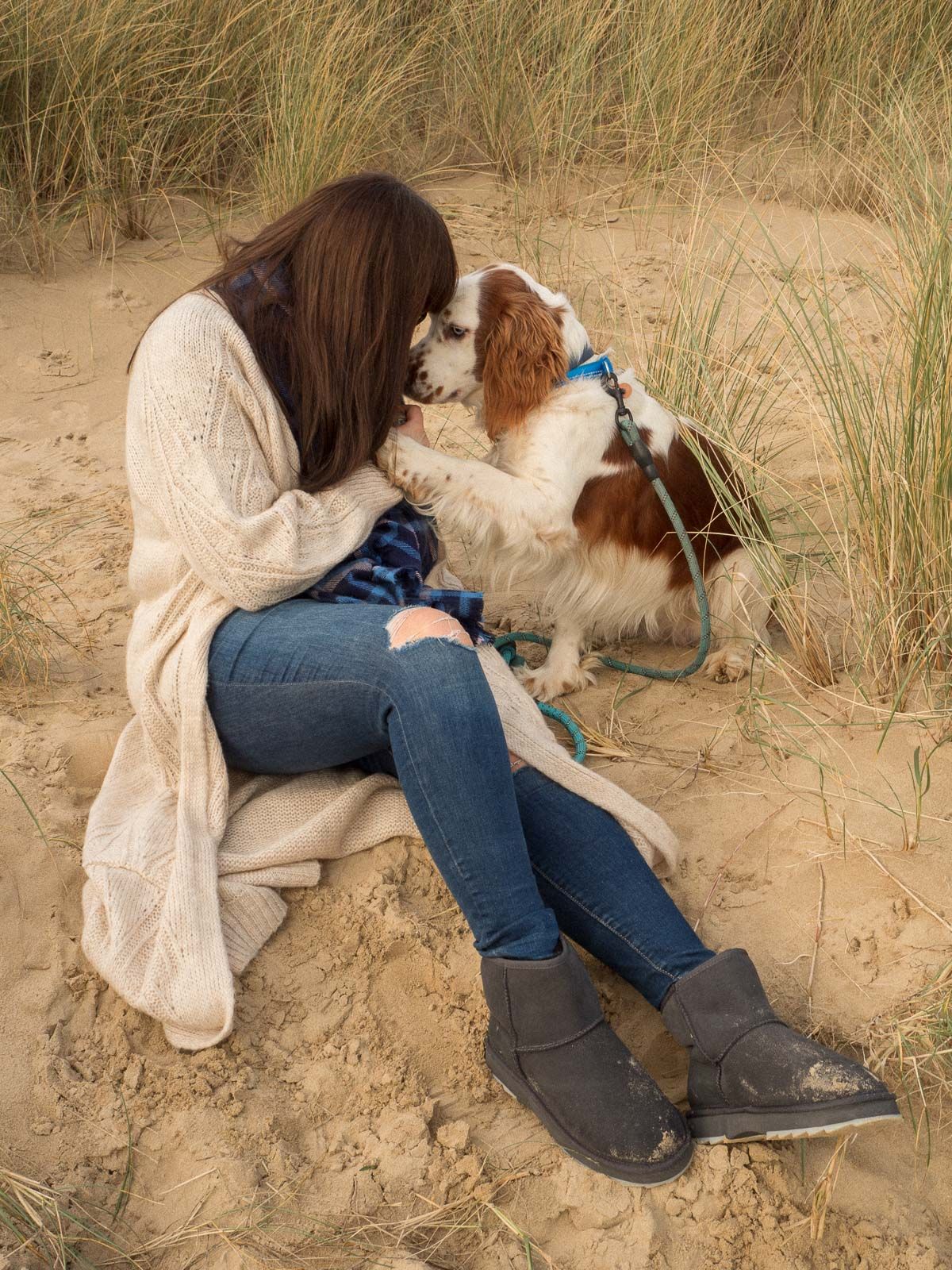}&
     \includegraphics[scale=0.073]{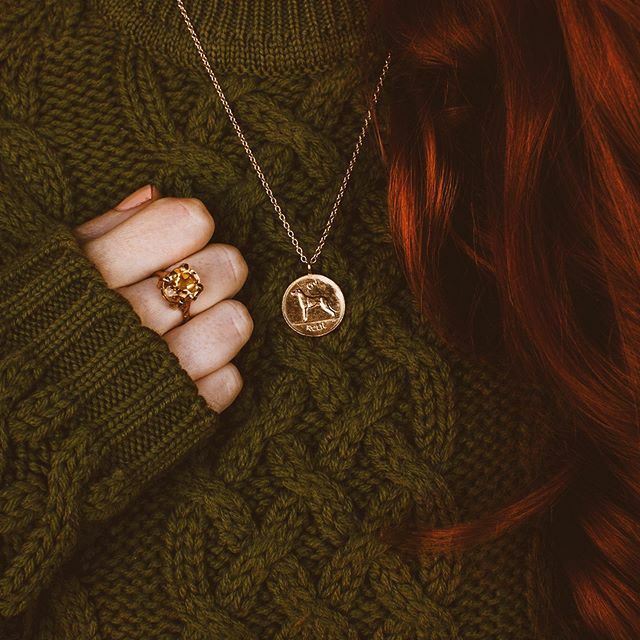} &  \includegraphics[scale=0.057]{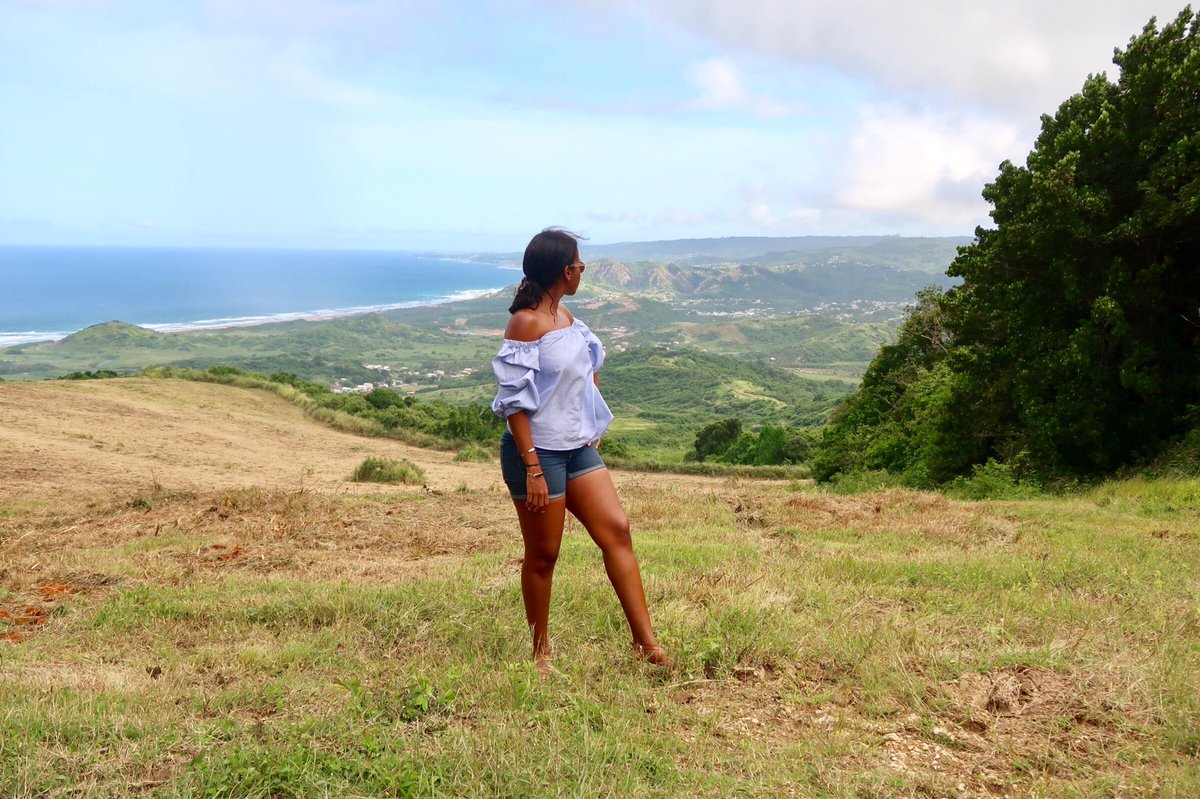}
     \\ 
     \begin{tabular}[l]{@{}l@{}} Combat the cold\\ weather with these\\ incredible @USER\\ sheepskin boots
    \end{tabular}  &
     \begin{tabular}[l]{@{}l@{}} chunky knits and dainty\\ jewels. This is my favor-\\ite vintage sweater\\ \#lovechupi \end{tabular}  & 
     \begin{tabular}[l]{@{}l@{}} Cherry tree hill is hands\\down the best view in\\ \#Barbados.\\ \#VisitBarbados \end{tabular}
    \\
    \hline
    \textbf{Actual:} C  &\textbf{Actual:} C & \textbf{Actual}: C \\
    \textbf{BERTweet:} C &\textbf{BERTweet:} NC & \textbf{BERTweet:} NC \\
    \textbf{ViT-BERTweet-Att:} NC&\textbf{ViT-BERTweet-Att:} C & \textbf{ViT-BERTweet-Att:} NC
        \\
        %\hline 
    \end{tabular}
%    }
    \caption{Examples of classifications of BERTweet and ViT-BERTweet-Att.}
    %\caption{Example of undisclosed \emph{commercial} post discovered by ViT-BERTweet-Att.}
    \label{fig:examples}
\end{figure}

\paragraph{Multimodal modeling errors.} The most common error when distinguishing \emph{commercial} posts ($60\%$) by our multimodal model, ViT-BERTweet-Att, corresponds to cases where the post includes a standard natural or personal photo, rather than an image depicting products, as is more common in influencer \emph{commercial} content \cite{kim2021discovering} and conventional online advertising \cite{al2022metadiscourse}. Figure \ref{fig:examples} Post A depicts a post incorrectly labeled as \emph{non-commercial} by ViT-BERTweet-Att and correclty classified by BERTweet.
%(see \emph{commercial} example in Figure \ref{fig:sampleLabels})

\paragraph{Multimodal modeling captures context beyond keyword-matching.} To analyze if multimodal modeling improves over weak labels, we apply the keyword-based weak labeling approach\footnote{Using the text before removing commercial keywords.} to the test sets (see Sec. \ref{sec:datalbls}). We find that $20\%$ and $80\%$ of the weak labeling errors in the text-image and the text-only test sets respectively, are correctly classified by ViT-BERTweet-Att. This suggests that our multimodal model, ViT-BERTweet-Att captures stylistic differences and visual information relevant to identify \emph{commercial} posts beyond keyword-matching. Indeed, most of the errors ($85\%$) in both text-image and text-only posts are false positives (i.e., true label is \emph{non-commercial}) and are misslabeled as \emph{commercial} as they contain one of the keywords, although they are used in a different context. For example: \textit{Just seen that Pepsi ad...awkward.}

\paragraph{Multimodal modeling aids in the discovery of undisclosed commercial posts} Using ViT-BERTweet-Att we found undisclosed \emph{commercial} posts ($15\%$) in text-image posts such as the one depicted in Figure \ref{fig:sampleLabels} (\emph{commercial}) and Figure \ref{fig:examples} Post B, as well as in text-only posts such as the next example: \textit{if you love @USER pro-collagen then you might like the new ultra smart line}.

\paragraph{Challenging cases for text and multimodal models.}
We observe cases that remain challenging for both multimodal and text-only models. Previous work in influencer commercial content on Instagram \citep{zarei2020characterising} highlights the difficulty of identifying commercial influencer posts promoting products given the use of \textit{native advertising} \citep{chia2012welcome}. However, we find that the most common error ($20\%$) when identifying \emph{commercial} posts (in both text-image and text-only posts), are those that rather than promoting products, they describe their ``personal" experiences, particularly while traveling, in both text and image as shown in Figure \ref{fig:examples} Post C. These \emph{commercial} posts are difficult to identify as they do not include any specific brand mention or product name and are accompanied by standard traveling images also common in \emph{non-commercial} posts \cite{oliveira2020people}.

\section{Conclusion}

We introduced a novel dataset of multimodal influencer content consisting of tweets labeled as \emph{commercial} or \emph{non-commercial}. This is the first dataset to include high quality annotated posts by experts in advertising regulation. %In this work, we also extend the list of \emph{commercial} keywords and show that our weak labeled training data is of good quality enabling training a range of unimodal and multimodal models for identifying influencer \emph{commercial} content. 
We conducted an extensive empirical study including vision, language and multimodal approaches as well as LLM prompting. Our results show that our proposed cross-attention approach to combine text and images, %can identify \emph{commercial} content with high performance in text-only and text-image posts, 
outperforms state-of-the-art multimodal models. Our new dataset can enable further studies on automatically detecting influencer hidden advertising as well as studies in computational linguistics \citep{sim-etal-2016-friends,sanchez-villegas-etal-2020-point,10.7717/peerj-cs.325,jin-etal-2022-automatic,ao-etal-2022-combining} for analysis of commercial language characteristics on a large scale. %work complements studies on commercial content and can contribute to promote ethical practices in the influencer marketing ecosystem. 
Future work includes modeling influencer content in multilingual settings.

\section*{Limitations}
We experimented using only data in English. Influencer advertising strategies could differ across cultures and languages. We plan to address this research direction in future work. We have also presented the main limitations of our best performing model in Section~\ref{sec:analysis}. %The majority of our human annotators are female. However, the emphasis in the annotation process has been on the understanding of market practices in the light of legal frameworks, which mitigates any potential gender imbalance in the annotator pool. Annotator details can be found in Appx. \ref{appdx:annotation}. %We have received advice from a national consumer protection authority to make sure our methodological approach relates to the public policy objectives of consumer disclosures. 

\section*{Ethics Statement}
Our work complies with Twitter data policy for research.\footnote{See: \url{https://developer.twitter.com/en/developer-terms/agreement-and-policy}} Tweets were retrieved in August 2021. We received approval from our University Research Ethics Committee.

\section*{Acknowledgments}
DSV and NA are supported by the Leverhulme Trust under Grant Number: RPG\#2020\#148. NA is also supported by ESRC (ES/T012714/1). DSV is also supported by the Centre for Doctoral Training in Speech and Language Technologies (SLT) and their Applications funded by the UK Research and Innovation grant EP/S023062/1. CG is supported by the ERC Starting Grant research project HUMANads (ERC-2021-StG No 101041824) and the Spinoza grant of the Dutch Research Council (NWO), awarded in 2021 to José van Dijck, Professor of Media and Digital Society at Utrecht University. We would also like to thank the members of the Competition and Markets Authority in the UK who contributed to the enhancement of the list of terms for our initial keyword-based strategy (refer to Section \ref{sec:datalbls}). Additionally, we extend our gratitude to the annotators who actively participated in our human annotation labeling task. We would like to thank Mali Jin, Yida Mu, Katerina Margatina, Constantinos Karouzos, Panayiotis Karachristou and all reviewers for their valuable feedback.

\bibliography{anthology,sponsored}

\begin{thebibliography}{71}
\expandafter\ifx\csname natexlab\endcsname\relax\def\natexlab#1{#1}\fi

\bibitem[{Al-Subhi(2022)}]{al2022metadiscourse}
Aisha~Saadi Al-Subhi. 2022.
\newblock Metadiscourse in online advertising: Exploring linguistic and visual metadiscourse in social media advertisements.
\newblock \emph{Journal of Pragmatics}, 187:24--40.

\bibitem[{Alassani and G{\"o}retz(2019)}]{alassani2019product}
Rachidatou Alassani and Julia G{\"o}retz. 2019.
\newblock Product placements by micro and macro influencers on instagram.
\newblock In \emph{International conference on human-computer interaction}, pages 251--267. Springer.

\bibitem[{Anderson et~al.(2018)Anderson, He, Buehler, Teney, Johnson, Gould, and Zhang}]{anderson2018bottom}
Peter Anderson, Xiaodong He, Chris Buehler, Damien Teney, Mark Johnson, Stephen Gould, and Lei Zhang. 2018.
\newblock Bottom-up and top-down attention for image captioning and visual question answering.
\newblock In \emph{Proceedings of the IEEE conference on computer vision and pattern recognition}, pages 6077--6086.

\bibitem[{Ao et~al.(2022)Ao, Sanchez~Villegas, Preotiuc-Pietro, and Aletras}]{ao-etal-2022-combining}
Xiao Ao, Danae Sanchez~Villegas, Daniel Preotiuc-Pietro, and Nikolaos Aletras. 2022.
\newblock \href {https://doi.org/10.18653/v1/2022.naacl-main.131} {Combining humor and sarcasm for improving political parody detection}.
\newblock In \emph{Proceedings of the 2022 Conference of the North American Chapter of the Association for Computational Linguistics: Human Language Technologies}, pages 1800--1807, Seattle, United States. Association for Computational Linguistics.

\bibitem[{Artstein and Poesio(2008)}]{artstein2008inter}
Ron Artstein and Massimo Poesio. 2008.
\newblock Inter-coder agreement for computational linguistics.
\newblock \emph{Computational linguistics}, 34(4):555--596.

\bibitem[{Boerman and van Reijmersdal(2016)}]{boerman2016informing}
Sophie~C Boerman and Eva~A van Reijmersdal. 2016.
\newblock Informing consumers about “hidden” advertising: A literature review of the effects of disclosing sponsored content.
\newblock \emph{Advertising in new formats and media}.

\bibitem[{Brown and Hayes(2008)}]{brown2008influencer}
Duncan Brown and Nick Hayes. 2008.
\newblock \emph{Influencer marketing}.
\newblock Routledge.

\bibitem[{Brown et~al.(2020)Brown, Mann, Ryder, Subbiah, Kaplan, Dhariwal, Neelakantan, Shyam, Sastry, Askell et~al.}]{brown2020language}
Tom Brown, Benjamin Mann, Nick Ryder, Melanie Subbiah, Jared~D Kaplan, Prafulla Dhariwal, Arvind Neelakantan, Pranav Shyam, Girish Sastry, Amanda Askell, et~al. 2020.
\newblock Language models are few-shot learners.
\newblock \emph{Advances in neural information processing systems}, 33:1877--1901.

\bibitem[{Chia(2012)}]{chia2012welcome}
Aleena Chia. 2012.
\newblock Welcome to me-mart: The politics of user-generated content in personal blogs.
\newblock \emph{American Behavioral Scientist}, 56(4):421--438.

\bibitem[{Chung et~al.(2022)Chung, Hou, Longpre, Zoph, Tay, Fedus, Li, Wang, Dehghani, Brahma et~al.}]{chung2022scaling}
Hyung~Won Chung, Le~Hou, Shayne Longpre, Barret Zoph, Yi~Tay, William Fedus, Eric Li, Xuezhi Wang, Mostafa Dehghani, Siddhartha Brahma, et~al. 2022.
\newblock Scaling instruction-finetuned language models.
\newblock \emph{arXiv preprint arXiv:2210.11416}.

\bibitem[{CMA(2020)}]{cma}
Competition {and} Markets~Authority CMA. 2020.
\newblock Influencers’ guide to making clear that ads are ads.
\newblock \url{https://www.ftc.gov/system/files/documents/plain-language/1001a-influencer-guide-508_1.pdf}.

\bibitem[{Cohen(1960)}]{cohen1960coefficient}
Jacob Cohen. 1960.
\newblock A coefficient of agreement for nominal scales.
\newblock \emph{Educational and psychological measurement}, 20(1):37--46.

\bibitem[{De~Gregorio and Goanta(2020)}]{de2020influencer}
Giovanni De~Gregorio and Catalina Goanta. 2020.
\newblock The influencer republic: Monetizing political speech on social media.
\newblock \emph{Available at SSRN}.

\bibitem[{Devlin et~al.(2019)Devlin, Chang, Lee, and Toutanova}]{devlin-etal-2019-bert}
Jacob Devlin, Ming-Wei Chang, Kenton Lee, and Kristina Toutanova. 2019.
\newblock \href {https://doi.org/10.18653/v1/N19-1423} {{BERT}: Pre-training of deep bidirectional transformers for language understanding}.
\newblock In \emph{Proceedings of the 2019 Conference of the North {A}merican Chapter of the Association for Computational Linguistics: Human Language Technologies, Volume 1 (Long and Short Papers)}, pages 4171--4186, Minneapolis, Minnesota. Association for Computational Linguistics.

\bibitem[{Dosovitskiy et~al.(2020)Dosovitskiy, Beyer, Kolesnikov, Weissenborn, Zhai, Unterthiner, Dehghani, Minderer, Heigold, Gelly et~al.}]{dosovitskiy2020image}
Alexey Dosovitskiy, Lucas Beyer, Alexander Kolesnikov, Dirk Weissenborn, Xiaohua Zhai, Thomas Unterthiner, Mostafa Dehghani, Matthias Minderer, Georg Heigold, Sylvain Gelly, et~al. 2020.
\newblock An image is worth 16x16 words: Transformers for image recognition at scale.
\newblock \emph{arXiv preprint arXiv:2010.11929}.

\bibitem[{Ducato(2020)}]{ducato2020one}
Rossana Ducato. 2020.
\newblock One hashtag to rule them all? mandated disclosures and design duties in influencer marketing practices.
\newblock In \emph{The Regulation of Social Media Influencers}. Edward Elgar Publishing.

\bibitem[{Ershov and Mitchell(2020)}]{ershov2020effects}
Daniel Ershov and Matthew Mitchell. 2020.
\newblock The effects of influencer advertising disclosure regulations: Evidence from instagram.
\newblock In \emph{Proceedings of the 21st ACM Conference on Economics and Computation}, pages 73--74.

\bibitem[{Evans et~al.(2017)Evans, Phua, Lim, and Jun}]{evans2017disclosing}
Nathaniel~J Evans, Joe Phua, Jay Lim, and Hyoyeun Jun. 2017.
\newblock Disclosing instagram influencer advertising: The effects of disclosure language on advertising recognition, attitudes, and behavioral intent.
\newblock \emph{Journal of interactive advertising}, 17(2):138--149.

\bibitem[{Fang and Wang(2022)}]{fang2022using}
Xing Fang and Tianfu Wang. 2022.
\newblock Using natural language processing to identify effective influencers.
\newblock \emph{International Journal of Market Research}, 64(5):611--629.

\bibitem[{Fernandes et~al.(2022)Fernandes, Rodrigues, and Shetty}]{10059732}
Roshan Fernandes, Anisha~P Rodrigues, and Bhuvaneshwari Shetty. 2022.
\newblock \href {https://doi.org/10.1109/AIDE57180.2022.10059732} {Influencers analysis from social media data}.
\newblock In \emph{2022 International Conference on Artificial Intelligence and Data Engineering (AIDE)}, pages 217--222.

\bibitem[{FTC(2019)}]{ftc}
Federal Trade~Commission FTC. 2019.
\newblock Disclosures 101 for social media influencers.
\newblock \url{https://www.ftc.gov/system/files/documents/plain-language/1001a-influencer-guide-508_1.pdf}.

\bibitem[{Gilbert et~al.(2013)Gilbert, Bakhshi, Chang, and Terveen}]{gilbert2013need}
Eric Gilbert, Saeideh Bakhshi, Shuo Chang, and Loren Terveen. 2013.
\newblock " i need to try this"? a statistical overview of pinterest.
\newblock In \emph{Proceedings of the SIGCHI conference on human factors in computing systems}, pages 2427--2436.

\bibitem[{Gross and Wangenheim(2018)}]{gross2018big}
Jana Gross and Florian~V Wangenheim. 2018.
\newblock The big four of influencer marketing. a typology of influencers.
\newblock \emph{Marketing Review St. Gallen}, 2:30--38.

\bibitem[{Han et~al.(2021)Han, Chen, Jin, Xu, Yang, Kumar, Zhao, Sundaram, and Kumar}]{10.1145/3449227}
Jinda Han, Qinglin Chen, Xilun Jin, Weikai Xu, Wanxian Yang, Suhansanu Kumar, Li~Zhao, Hari Sundaram, and Ranjitha Kumar. 2021.
\newblock \href {https://doi.org/10.1145/3449227} {Fitnet: Identifying fashion influencers on twitter}.
\newblock \emph{Proc. ACM Hum.-Comput. Interact.}, 5(CSCW1).

\bibitem[{He et~al.(2016)He, Zhang, Ren, and Sun}]{he2016deep}
Kaiming He, Xiangyu Zhang, Shaoqing Ren, and Jian Sun. 2016.
\newblock Deep residual learning for image recognition.
\newblock In \emph{Proceedings of the IEEE conference on computer vision and pattern recognition}, pages 770--778.

\bibitem[{Hessel and Lee(2020)}]{hessel-lee-2020-multimodal}
Jack Hessel and Lillian Lee. 2020.
\newblock \href {https://doi.org/10.18653/v1/2020.emnlp-main.62} {Does my multimodal model learn cross-modal interactions? it{'}s harder to tell than you might think!}
\newblock In \emph{Proceedings of the 2020 Conference on Empirical Methods in Natural Language Processing (EMNLP)}, pages 861--877, Online. Association for Computational Linguistics.

\bibitem[{Hochreiter and Schmidhuber(1997)}]{hochreiter1997long}
Sepp Hochreiter and J{\"u}rgen Schmidhuber. 1997.
\newblock Long short-term memory.
\newblock \emph{Neural computation}, 9(8):1735--1780.

\bibitem[{Huang et~al.(2014)Huang, Kornfield, Szczypka, and Emery}]{huang2014cross}
Jidong Huang, Rachel Kornfield, Glen Szczypka, and Sherry~L Emery. 2014.
\newblock A cross-sectional examination of marketing of electronic cigarettes on twitter.
\newblock \emph{Tobacco control}, 23(suppl 3):iii26--iii30.

\bibitem[{Jarrar et~al.(2020)Jarrar, Awobamise, and Aderibigbe}]{jarrar2020effectiveness}
Yosra Jarrar, Ayodeji~Olalekan Awobamise, and Adebola~Adewunmi Aderibigbe. 2020.
\newblock Effectiveness of influencer marketing vs social media sponsored advertising.
\newblock \emph{Utopia y Praxis Latinoamericana}, 25(12):40--54.

\bibitem[{Jin et~al.(2022)Jin, Preotiuc-Pietro, Do{\u{g}}ru{\"o}z, and Aletras}]{jin-etal-2022-automatic}
Mali Jin, Daniel Preotiuc-Pietro, A.~Seza Do{\u{g}}ru{\"o}z, and Nikolaos Aletras. 2022.
\newblock \href {https://doi.org/10.18653/v1/2022.acl-long.273} {Automatic identification and classification of bragging in social media}.
\newblock In \emph{Proceedings of the 60th Annual Meeting of the Association for Computational Linguistics (Volume 1: Long Papers)}, pages 3945--3959, Dublin, Ireland. Association for Computational Linguistics.

\bibitem[{Kay et~al.(2020)Kay, Mulcahy, and Parkinson}]{kay2020less}
Samantha Kay, Rory Mulcahy, and Joy Parkinson. 2020.
\newblock When less is more: the impact of macro and micro social media influencers’ disclosure.
\newblock \emph{Journal of Marketing Management}, 36(3-4):248--278.

\bibitem[{Keller and Berry(2003)}]{keller2003influentials}
Edward Keller and Jonathan Berry. 2003.
\newblock \emph{The influentials: One American in ten tells the other nine how to vote, where to eat, and what to buy}.
\newblock Simon and Schuster.

\bibitem[{Kiela et~al.(2019)Kiela, Bhooshan, Firooz, Perez, and Testuggine}]{kiela2019supervised}
Douwe Kiela, Suvrat Bhooshan, Hamed Firooz, Ethan Perez, and Davide Testuggine. 2019.
\newblock Supervised multimodal bitransformers for classifying images and text.
\newblock \emph{arXiv preprint arXiv:1909.02950}.

\bibitem[{Kim et~al.(2021{\natexlab{a}})Kim, Chen, Jiang, Han, and Wang}]{kim2021evaluating}
Seungbae Kim, Xiusi Chen, Jyun-Yu Jiang, Jinyoung Han, and Wei Wang. 2021{\natexlab{a}}.
\newblock Evaluating audience loyalty and authenticity in influencer marketing via multi-task multi-relational learning.
\newblock In \emph{Proceedings of the International AAAI Conference on Web and Social Media}, volume~15, pages 278--289.

\bibitem[{Kim et~al.(2017)Kim, Han, Yoo, and Gerla}]{kim2017social}
Seungbae Kim, Jinyoung Han, Seunghyun Yoo, and Mario Gerla. 2017.
\newblock How are social influencers connected in instagram?
\newblock In \emph{International Conference on Social Informatics}, pages 257--264. Springer.

\bibitem[{Kim et~al.(2020)Kim, Jiang, Nakada, Han, and Wang}]{kim2020multimodal}
Seungbae Kim, Jyun-Yu Jiang, Masaki Nakada, Jinyoung Han, and Wei Wang. 2020.
\newblock Multimodal post attentive profiling for influencer marketing.
\newblock In \emph{Proceedings of The Web Conference 2020}, pages 2878--2884.

\bibitem[{Kim et~al.(2021{\natexlab{b}})Kim, Jiang, and Wang}]{kim2021discovering}
Seungbae Kim, Jyun-Yu Jiang, and Wei Wang. 2021{\natexlab{b}}.
\newblock Discovering undisclosed paid partnership on social media via aspect-attentive sponsored post learning.
\newblock In \emph{Proceedings of the 14th ACM International Conference on Web Search and Data Mining}, pages 319--327.

\bibitem[{Kim et~al.(2021{\natexlab{c}})Kim, Son, and Kim}]{kim2021vilt}
Wonjae Kim, Bokyung Son, and Ildoo Kim. 2021{\natexlab{c}}.
\newblock Vilt: Vision-and-language transformer without convolution or region supervision.
\newblock In \emph{International Conference on Machine Learning}, pages 5583--5594. PMLR.

\bibitem[{King and Zeng(2001)}]{king2001logistic}
Gary King and Langche Zeng. 2001.
\newblock Logistic regression in rare events data.
\newblock \emph{Political analysis}, 9(2):137--163.

\bibitem[{Kingma and Ba(2014)}]{kingma2014adam}
Diederik~P Kingma and Jimmy Ba. 2014.
\newblock Adam: A method for stochastic optimization.
\newblock \emph{arXiv preprint arXiv:1412.6980}.

\bibitem[{Lagr{\'e}e et~al.(2018)Lagr{\'e}e, Capp{\'e}, Cautis, and Maniu}]{lagree2018algorithms}
Paul Lagr{\'e}e, Olivier Capp{\'e}, Bogdan Cautis, and Silviu Maniu. 2018.
\newblock Algorithms for online influencer marketing.
\newblock \emph{ACM Transactions on Knowledge Discovery from Data (TKDD)}, 13(1):1--30.

\bibitem[{Lee et~al.(2022)Lee, Sudarshan, Sussman, Bright, and Eastin}]{lee2022consumers}
Jung~Ah Lee, Sabitha Sudarshan, Kristen~L Sussman, Laura~F Bright, and Matthew~S Eastin. 2022.
\newblock Why are consumers following social media influencers on instagram? exploration of consumers’ motives for following influencers and the role of materialism.
\newblock \emph{International Journal of Advertising}, 41(1):78--100.

\bibitem[{Leerssen et~al.(2019)Leerssen, Ausloos, Zarouali, Helberger, and de~Vreese}]{Leerssen2019}
Paddy Leerssen, Jef Ausloos, Brahim Zarouali, Natali Helberger, and Claes~H. de~Vreese. 2019.
\newblock \href {https://doi.org/10.14763/2019.4.1421} {Platform ad archives: promises and pitfalls}.
\newblock \emph{Internet Policy Review}, 8(4).

\bibitem[{Lou et~al.(2019)Lou, Tan, and Chen}]{lou2019investigating}
Chen Lou, Sang-Sang Tan, and Xiaoyu Chen. 2019.
\newblock Investigating consumer engagement with influencer-vs. brand-promoted ads: The roles of source and disclosure.
\newblock \emph{Journal of Interactive Advertising}, 19(3):169--186.

\bibitem[{Lu et~al.(2019)Lu, Batra, Parikh, and Lee}]{NEURIPS2019_c74d97b0}
Jiasen Lu, Dhruv Batra, Devi Parikh, and Stefan Lee. 2019.
\newblock \href {https://proceedings.neurips.cc/paper/2019/file/c74d97b01eae257e44aa9d5bade97baf-Paper.pdf} {Vilbert: Pretraining task-agnostic visiolinguistic representations for vision-and-language tasks}.
\newblock In \emph{Advances in Neural Information Processing Systems}, volume~32. Curran Associates, Inc.

\bibitem[{Ma et~al.(2021)Ma, Shen, Yoshikawa, Iwakura, Beck, and Baldwin}]{ma-etal-2021-effectiveness}
Chunpeng Ma, Aili Shen, Hiyori Yoshikawa, Tomoya Iwakura, Daniel Beck, and Timothy Baldwin. 2021.
\newblock \href {https://doi.org/10.18653/v1/2021.eacl-main.4} {On the (in)effectiveness of images for text classification}.
\newblock In \emph{Proceedings of the 16th Conference of the European Chapter of the Association for Computational Linguistics: Main Volume}, pages 42--48, Online. Association for Computational Linguistics.

\bibitem[{Mathur et~al.(2018)Mathur, Narayanan, and Chetty}]{mathur2018endorsements}
Arunesh Mathur, Arvind Narayanan, and Marshini Chetty. 2018.
\newblock Endorsements on social media: An empirical study of affiliate marketing disclosures on youtube and pinterest.
\newblock \emph{Proceedings of the ACM on Human-Computer Interaction}, 2(CSCW):1--26.

\bibitem[{Mazloom et~al.(2016)Mazloom, Rietveld, Rudinac, Worring, and Van~Dolen}]{mazloom2016multimodal}
Masoud Mazloom, Robert Rietveld, Stevan Rudinac, Marcel Worring, and Willemijn Van~Dolen. 2016.
\newblock Multimodal popularity prediction of brand-related social media posts.
\newblock In \emph{Proceedings of the 24th ACM international conference on Multimedia}, pages 197--201.

\bibitem[{Mu and Aletras(2020)}]{10.7717/peerj-cs.325}
Yida Mu and Nikolaos Aletras. 2020.
\newblock \href {https://doi.org/10.7717/peerj-cs.325} {Identifying twitter users who repost unreliable news sources with linguistic information}.
\newblock \emph{PeerJ Computer Science}, 6:e325.

\bibitem[{Nandagiri and Philip(2018)}]{nandagiri2018impact}
Vaibhavi Nandagiri and Leena Philip. 2018.
\newblock Impact of influencers from instagram and youtube on their followers.
\newblock \emph{International Journal of Multidisciplinary Research and Modern Education}, 4(1):61--65.

\bibitem[{Nguyen et~al.(2020)Nguyen, Vu, and Tuan~Nguyen}]{nguyen-etal-2020-bertweet}
Dat~Quoc Nguyen, Thanh Vu, and Anh Tuan~Nguyen. 2020.
\newblock \href {https://doi.org/10.18653/v1/2020.emnlp-demos.2} {{BERT}weet: A pre-trained language model for {E}nglish tweets}.
\newblock In \emph{Proceedings of the 2020 Conference on Empirical Methods in Natural Language Processing: System Demonstrations}, pages 9--14, Online. Association for Computational Linguistics.

\bibitem[{Oliveira et~al.(2020)Oliveira, Araujo, and Tam}]{oliveira2020people}
Tiago Oliveira, Benedita Araujo, and Carlos Tam. 2020.
\newblock Why do people share their travel experiences on social media?
\newblock \emph{Tourism Management}, 78:104041.

\bibitem[{Pennington et~al.(2014)Pennington, Socher, and Manning}]{pennington-etal-2014-glove}
Jeffrey Pennington, Richard Socher, and Christopher Manning. 2014.
\newblock \href {https://doi.org/10.3115/v1/D14-1162} {{G}lo{V}e: Global vectors for word representation}.
\newblock In \emph{Proceedings of the 2014 Conference on Empirical Methods in Natural Language Processing ({EMNLP})}, pages 1532--1543, Doha, Qatar. Association for Computational Linguistics.

\bibitem[{Ren et~al.(2016)Ren, He, Girshick, and Sun}]{ren2016faster}
Shaoqing Ren, Kaiming He, Ross Girshick, and Jian Sun. 2016.
\newblock Faster {R}-{C}{N}{N}: towards real-time object detection with region proposal networks.
\newblock \emph{IEEE transactions on pattern analysis and machine intelligence}, 39(6):1137--1149.

\bibitem[{Russakovsky et~al.(2015)Russakovsky, Deng, Su, Krause, Satheesh, Ma, Huang, Karpathy, Khosla, Bernstein et~al.}]{russakovsky2015imagenet}
Olga Russakovsky, Jia Deng, Hao Su, Jonathan Krause, Sanjeev Satheesh, Sean Ma, Zhiheng Huang, Andrej Karpathy, Aditya Khosla, Michael Bernstein, et~al. 2015.
\newblock Imagenet large scale visual recognition challenge.
\newblock \emph{International journal of computer vision}, 115:211--252.

\bibitem[{S{\'a}nchez~Villegas and Aletras(2021)}]{sanchez-villegas-aletras-2021-point}
Danae S{\'a}nchez~Villegas and Nikolaos Aletras. 2021.
\newblock \href {https://doi.org/10.18653/v1/2021.emnlp-main.614} {Point-of-interest type prediction using text and images}.
\newblock In \emph{Proceedings of the 2021 Conference on Empirical Methods in Natural Language Processing}, pages 7785--7797, Online and Punta Cana, Dominican Republic. Association for Computational Linguistics.

\bibitem[{S{\'a}nchez~Villegas et~al.(2021)S{\'a}nchez~Villegas, Mokaram, and Aletras}]{sanchez-villegas-etal-2021-analyzing}
Danae S{\'a}nchez~Villegas, Saeid Mokaram, and Nikolaos Aletras. 2021.
\newblock \href {https://doi.org/10.18653/v1/2021.findings-acl.321} {Analyzing online political advertisements}.
\newblock In \emph{Findings of the Association for Computational Linguistics: ACL-IJCNLP 2021}, pages 3669--3680, Online. Association for Computational Linguistics.

\bibitem[{S{\'a}nchez~Villegas et~al.(2020)S{\'a}nchez~Villegas, Preotiuc-Pietro, and Aletras}]{sanchez-villegas-etal-2020-point}
Danae S{\'a}nchez~Villegas, Daniel Preotiuc-Pietro, and Nikolaos Aletras. 2020.
\newblock \href {https://aclanthology.org/2020.aacl-main.80} {Point-of-interest type inference from social media text}.
\newblock In \emph{Proceedings of the 1st Conference of the Asia-Pacific Chapter of the Association for Computational Linguistics and the 10th International Joint Conference on Natural Language Processing}, pages 804--810, Suzhou, China. Association for Computational Linguistics.

\bibitem[{Schwartz et~al.(2017)Schwartz, Giorgi, Sap, Crutchley, Ungar, and Eichstaedt}]{schwartz-etal-2017-dlatk}
H.~Andrew Schwartz, Salvatore Giorgi, Maarten Sap, Patrick Crutchley, Lyle Ungar, and Johannes Eichstaedt. 2017.
\newblock \href {https://doi.org/10.18653/v1/D17-2010} {{DLATK}: Differential language analysis {T}ool{K}it}.
\newblock In \emph{Proceedings of the 2017 Conference on Empirical Methods in Natural Language Processing: System Demonstrations}, pages 55--60, Copenhagen, Denmark. Association for Computational Linguistics.

\bibitem[{Sim et~al.(2016)Sim, Routledge, and Smith}]{sim-etal-2016-friends}
Yanchuan Sim, Bryan Routledge, and Noah~A. Smith. 2016.
\newblock \href {https://doi.org/10.18653/v1/D16-1178} {{F}riends with motives: Using text to infer influence on {SCOTUS}}.
\newblock In \emph{Proceedings of the 2016 Conference on Empirical Methods in Natural Language Processing}, pages 1724--1733, Austin, Texas. Association for Computational Linguistics.

\bibitem[{Tan and Bansal(2019)}]{tan-bansal-2019-lxmert}
Hao Tan and Mohit Bansal. 2019.
\newblock \href {https://doi.org/10.18653/v1/D19-1514} {{LXMERT}: Learning cross-modality encoder representations from transformers}.
\newblock In \emph{Proceedings of the 2019 Conference on Empirical Methods in Natural Language Processing and the 9th International Joint Conference on Natural Language Processing (EMNLP-IJCNLP)}, pages 5100--5111, Hong Kong, China. Association for Computational Linguistics.

\bibitem[{Vaswani et~al.(2017)Vaswani, Shazeer, Parmar, Uszkoreit, Jones, Gomez, Kaiser, and Polosukhin}]{Vaswani2017}
Ashish Vaswani, Noam Shazeer, Niki Parmar, Jakob Uszkoreit, Llion Jones, Aidan~N Gomez, {\L}ukasz Kaiser, and Illia Polosukhin. 2017.
\newblock Attention is all you need.
\newblock In \emph{Advances in Neural Information Processing Systems}, pages 5998--6008.

\bibitem[{Vempala and Preo{\c{t}}iuc-Pietro(2019)}]{vempala-preotiuc-pietro-2019-categorizing}
Alakananda Vempala and Daniel Preo{\c{t}}iuc-Pietro. 2019.
\newblock \href {https://doi.org/10.18653/v1/P19-1272} {Categorizing and inferring the relationship between the text and image of {T}witter posts}.
\newblock In \emph{Proceedings of the 57th Annual Meeting of the Association for Computational Linguistics}, pages 2830--2840, Florence, Italy. Association for Computational Linguistics.

\bibitem[{Wang et~al.(2020)Wang, Li, Lyu, and King}]{wang-etal-2020-cross-media}
Yue Wang, Jing Li, Michael Lyu, and Irwin King. 2020.
\newblock \href {https://doi.org/10.18653/v1/2020.emnlp-main.268} {Cross-media keyphrase prediction: A unified framework with multi-modality multi-head attention and image wordings}.
\newblock In \emph{Proceedings of the 2020 Conference on Empirical Methods in Natural Language Processing (EMNLP)}, pages 3311--3324, Online. Association for Computational Linguistics.

\bibitem[{Weber et~al.(2021)Weber, Ludwig, Brodesser, and Gr{\"o}newald}]{weber2021s}
Philip Weber, Thomas Ludwig, Sabrina Brodesser, and Laura Gr{\"o}newald. 2021.
\newblock “it's a kind of art!”: Understanding food influencers as influential content creators.
\newblock In \emph{Proceedings of the 2021 CHI Conference on Human Factors in Computing Systems}, pages 1--14.

\bibitem[{Wojdynski(2016)}]{wojdynski2016native}
Bartosz~W Wojdynski. 2016.
\newblock Native advertising: Engagement, deception, and implications for theory.
\newblock \emph{The new advertising: Branding, content and consumer relationships in a data-driven social media era}, pages 203--236.

\bibitem[{Wojdynski et~al.(2018)Wojdynski, Evans, and Hoy}]{wojdynski2018measuring}
Bartosz~W Wojdynski, Nathaniel~J Evans, and Mariea~Grubbs Hoy. 2018.
\newblock Measuring sponsorship transparency in the age of native advertising.
\newblock \emph{Journal of Consumer Affairs}, 52(1):115--137.

\bibitem[{Wolf et~al.(2019)Wolf, Debut, Sanh, Chaumond, Delangue, Moi, Cistac, Rault, Louf, Funtowicz et~al.}]{wolf2019huggingface}
Thomas Wolf, Lysandre Debut, Victor Sanh, Julien Chaumond, Clement Delangue, Anthony Moi, Pierric Cistac, Tim Rault, R{\'e}mi Louf, Morgan Funtowicz, et~al. 2019.
\newblock Huggingface's transformers: State-of-the-art natural language processing.
\newblock \emph{arXiv preprint arXiv:1910.03771}.

\bibitem[{Yang et~al.(2019)Yang, Kim, and Sun}]{yang2019influencers}
Xiao Yang, Seungbae Kim, and Yizhou Sun. 2019.
\newblock How do influencers mention brands in social media? sponsorship prediction of instagram posts.
\newblock In \emph{Proceedings of the 2019 IEEE/ACM International Conference on Advances in Social Networks Analysis and Mining}, pages 101--104.

\bibitem[{Zarei et~al.(2020)Zarei, Ibosiola, Farahbakhsh, Gilani, Garimella, Crespi, and Tyson}]{zarei2020characterising}
Koosha Zarei, Damilola Ibosiola, Reza Farahbakhsh, Zafar Gilani, Kiran Garimella, No{\"e}l Crespi, and Gareth Tyson. 2020.
\newblock Characterising and detecting sponsored influencer posts on instagram.
\newblock In \emph{2020 IEEE/ACM International Conference on Advances in Social Networks Analysis and Mining (ASONAM)}, pages 327--331. IEEE.

\bibitem[{Zhou et~al.(2020)Zhou, Palangi, Zhang, Hu, Corso, and Gao}]{Zhou_Palangi_Zhang_Hu_Corso_Gao_2020}
Luowei Zhou, Hamid Palangi, Lei Zhang, Houdong Hu, Jason Corso, and Jianfeng Gao. 2020.
\newblock \href {https://doi.org/10.1609/aaai.v34i07.7005} {Unified {V}ision-{L}anguage {P}re-{T}raining for {I}mage {C}aptioning and {V}{Q}{A}}.
\newblock \emph{Proceedings of the AAAI Conference on Artificial Intelligence}, 34(07):13041--13049.

\end{thebibliography}
\bibliographystyle{acl_natbib}

\newpage
\appendix

\section{Influencer Marketing Keywords}
\label{sec:appendix_keywords}
We extract keywords strongly associated with influencer marketing from the guidelines provided by the Federal Trade Commission \citep{ftc} %\footnote{\url{https://www.ftc.gov/system/files/documents/plain-language/1001a-influencer-guide-508_1.pdf}} 
in the US, and the Advertisements Standards Authority and Competition and Markets Authority in the UK \citep{cma}. %\footnote{\url{https://www.asa.org.uk/static/9cc1fb3f-1288-405d-af3468ff18277299/INFLUENCERGuidanceupdatev6HR.pdf}} (see Sec. \ref{sec:datalbls}).
The keywords in these guidelines are based on regulatory standards for digital enforcement which are meant to create objective and transparent expectations regarding the disclosure of \textit{native advertising} on social media. Thus, our list of keywords include sponsorship disclosure terms that are relevant to different business models (i.e., market practices based on the obligations of the parties). A complete list of keywords is presented in Table \ref{tab:bmodels}.

\begin{table*}[]
\centering
\small
\begin{tabular}{|l|l|l|}
\hline
\multicolumn{1}{|c|}{\textbf{\begin{tabular}[c]{@{}c@{}} Type\end{tabular}}} &
\multicolumn{1}{c|}{\textbf{Description}} & \multicolumn{1}{c|}{\textbf{Commercial Keywords}} \\ \hline
\begin{tabular}[c]{@{}l@{}}Guidelines\end{tabular} &  \begin{tabular}[l]{@{}l@{}} Keywords retrieved from relevant guidelines \\Recommended and not recommended terms. \end{tabular}
& \begin{tabular}[c]{@{}l@{}} \#ad, ad, \#advert \#collab, collab, \#spon, \#sponsored,\\ spon, \#sp, sponsored, `thanks to'/ `funded by'/\\`supported by'/`in association with' @USER \end{tabular} \\ \hline
Endoresements & 
\begin{tabular}[l]{@{}l@{}}An influencer receives money to promote a\\ product or service.
\end{tabular}
&\#ambassador, ambassador \\ \hline
Barter &
\begin{tabular}[l]{@{}l@{}} Exchange of goods or services from a brand\\ or its representatives against an advertising\\ service offered by the influencer.
\end{tabular}
&\begin{tabular}[c]{@{}l@{}}\#gift, gift, \#giveaway, giveaway \\ unpaid sample \end{tabular} \\\hline
\begin{tabular}[c]{@{}l@{}}Affiliate\\ Marketing\end{tabular} & 
\begin{tabular}[l]{@{}l@{}} The influencer is paid a percentage of referral\\ sales, often identified through discount\\ codes.
\end{tabular}
&\begin{tabular}[c]{@{}l@{}}\#aff, aff, \#affiliate, affiliate, \\ discount  code\end{tabular} \\ \hline

\end{tabular}
\caption{Commercial keywords. @USER refers to an @-mention of a brand account.}
\label{tab:bmodels}
\end{table*}

\begin{table*}[t!]
\small\centering
\begin{tabular}{|l|c|l|}
\hline
\textbf{Dataset}                & \textbf{\begin{tabular}[c]{@{}l@{}}No. of \\Commercial\\ Keywords\end{tabular}} & \textbf{Commercial Keywords}                                                                                                                                                                                                                                                                                                                   \\ \hline
\citet{10.1145/3449227}         & 0                                                                        & -                                                                                                                                                                                                                                                                                                                                             \\ \hline
\citet{zarei2020characterising} & 7                                                                        & \begin{tabular}[c]{@{}l@{}}\#ad, \#advert, \#sponsored\, \#advertising,\\  \#giveaway, \#spon, \#sponsor\end{tabular}                                                                                                                                                                                                                        \\ \hline
\citet{yang2019influencers}     & 3                                                                        & \#ad, \#sponsored, \#paidAD                                                                                                                                                                                                                                                                                                                    \\ \hline
\citet{kim2021discovering}      & 3                                                                        & \#ad, \#sponsored, \#paidA                                                                                                                                                                                                                                                                                                                     \\ \hline
\citet{kim2020multimodal}       & 1                                                                        & \#ad                                                                                                                                                                                                                                                                                                                                           \\ \hline
MICD (Ours)                     & 26                                                                       & \begin{tabular}[c]{@{}l@{}}\#ad, ad, \#advert, \#sponsored, \#collab, \\ collab, spon, \#sp, sponsored, \#aff, aff,\\  `thanks to'/ `funded by'/, unpaid sample,\\ `supported by'/`in association with' @USER, \\ \#ambassador, ambassador, discount code\\  \#gift, gift, \#giveaway, giveaway, \#spon\\ \#affiliate, affiliate,\end{tabular} \\ \hline
\end{tabular}
\caption{Comparison of commercial keywords used in existing datasets and in ours (MICD)}
\label{tab:dataset-kwcomp}
\end{table*}

\section{Annotation Guidelines}
\label{appdx:annotation}
\begin{figure*}[t!]
\centering
\includegraphics[scale=0.29]{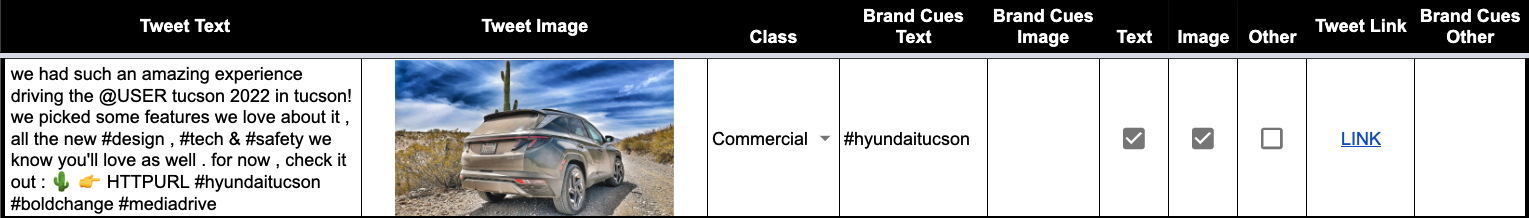}
\caption{Example of Annotation}
\label{fig:method}
\end{figure*}

\paragraph{Purpose of the study} This annotation effort is part of a study that aims to characterize and identify commercial content on Twitter. Commercial content is an umbrella term for communications that relate to commercial transactions, or in other words, content that is monetized. For influencers, that may entail various business models: 

\begin{itemize}
    \item Endorsements: an influencer receives money in order to promote a product or service.
    \item Affiliate marketing: the influencer is paid a percentage of referral sales, often identified through discount codes.
    \item Barter: exchange of goods or services from a brand or its representatives against an advertising service offered by the influencer.
    \item Direct selling: influencers can also choose to create their own products, branded products, and/or services, and link to their web shops. 
\end{itemize}

\paragraph{Task Description}

The task is to annotate whether a given influencer’s Twitter post is perceived to contain commercial content or not given only its text and image content (if available).
If annotators perceive that the tweet contains commercial content, then it should be annotated as commercial, otherwise as non-commercial. If it is not clear whether the Tweet is perceived to contain commercial content, it should be labeled as unclear. The details of each category are as follows:

\begin{itemize}
    \item Commercial: posts refer to any of the business models mentioned above. This category includes promoting or endorsing a brand or its products/services, a free loan of a product/service, a free product/service (whether requested or received out of the blue), or any other incentive. This can be noted by the use of terms or hashtags such as \#gifted, \#ad, @mentions of the brand, hashtags including the name of the brand and/or campaign slogans.
    \item Non-Commercial: Organic content such as personal ideas, personal comments and life updates, and that does not seem monetized through any of the business models mentioned above. 
    \item Unclear: This option should be chosen when it is not clear whether the Tweet contains commercial content or not (e.g., commenting about a brand without using hashtags or @mentioning the brand).
\end{itemize}

\paragraph{Instructions}
\begin{enumerate}
\item For each post, read the text, look at the image (if available), and select one of the categories (Commercial, Non-commercial, Unclear). 
\item If the post is annotated as Commercial, then in the ``Brand Cues" section write down the term(s) or hashtag(s) that support your decision such as: \#gifted, \#ad, @mentions, hashtags including the name of the brand and/or campaign slogans. Use the ``Brand Cues" column that corresponds to the location of them: ``Brand Cues Text" if the brand cues are found in the text and/or "Brand Cues Image" if they are located in the image.
Select the option(s) (Text, Image) used to make your annotation (e.g., if the brand cues are in the text then select Text, if the post was annotated as non-commercial choose the option that you looked at to make your decision). 
\item If the post was annotated as Unclear, then: select the ``Other" option and click on the Tweet Link. If you find any brand cues in the Tweet’s page, write them down in the column “Brand cues Other”. If it is still unclear whether the Tweet is commercial or not keep the label ``Unclear", otherwise select the appropriate label (Commercial/Non-Commercial). 
\end{enumerate}

\paragraph{Annotator Details}
All annotators were senior law school students (third year bachelor and masters level) who study comparative and international law. The students have a background in law, which entails a good grasp of consumer protection disclosures. In addition, their profiles were also particularly interesting for annotation since they had spent 6 months of their study being trained under an extracurricular Influencer Law Clinic honors programme. The training consisted in multidisciplinary workshops and hands-on research on influencer-related legal topics. The annotators come from a wide range of socio-economic backgrounds and are fluent in English. The majority of annotators are female. However, the emphasis in the annotation process has been on the understanding of market practices in the light of legal frameworks, which mitigates any potential gender imbalance in the annotator pool. All annotators expressed their written consent and were informed about how data would be used following ethics guidelines from our Institution.

\begin{figure}[t!]
\small
\centering
\includegraphics[scale=0.47]{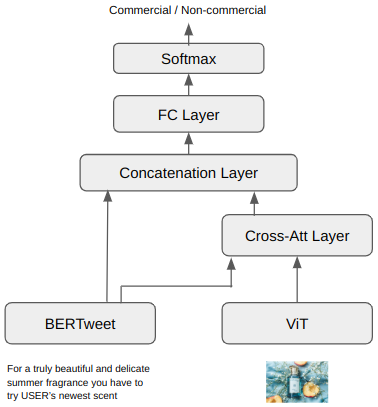}
\caption{ViT-BERTweet-Att model for detecting \emph{commercial} content. FC: fully-connected layer.}
\label{fig:model}
\end{figure}

\section{Predictive Performance}
\label{sec:appendix_res}
Table \ref{tab:results_macro_appx} and Table \ref{tab:res_txtonly_macro_appdx} present the macro-averaged results of \emph{commercial} content prediction. % including standard deviations over three runs using different random seeds. 

\section{Prompt Templates}
\label{appdx:prompting}

\subsection{Zero-shot Prompting}
For zero-shot prompting we use the following prompt: 
\begin{quote}
\small
\it
\noindent\textit{Label the next text as `commercial' or `not commercial'. Text: <TWEET>}.
\end{quote}

\noindent We map responses to the corresponding \emph{commercial} or \emph{non-commercial} class and report results for each model.

\subsection{Few-shot Prompting}
We experiment with few-shot prompting by appending four randomly selected training examples (two examples from each class) before each prompt. We run this three times with a different set examples. Table \ref{tab:results} shows average and standard deviation performance. The few-shot prompt follows the next template:

\begin{quote}
\small
\it
\noindent\textit{Label the next text as `commercial' or `not commercial'. Text: <TWEET-TRAIN> // <LABEL-TRAIN>}

\noindent\textit{Label the next text as `commercial' or `not commercial'. Text: <TWEET-TRAIN> // <LABEL-TRAIN>}

\noindent\textit{Label the next text as `commercial' or `not commercial'. Text: <TWEET-TRAIN> // <LABEL-TRAIN>}

\noindent\textit{Label the next text as `commercial' or `not commercial'. Text: <TWEET-TRAIN> // <LABEL-TRAIN>"}

\noindent\textit{Label the next text as `commercial' or `not commercial'. Text: <TWEET> //}

\end{quote}

\noindent\small <Label-TRAIN> \normalsize corresponds to the true label of the \small<TWEET-TRAIN> \normalsize 
training example (\emph{commercial} or \emph{non-commercial}), \small<TWEET> \normalsize refers to a testing example. We remove punctuation and spaces and map the output of each model (FLAN-T5 or GPT-3) to the corresponding label (\emph{commercial} or \emph{non-commercial}).

% \subsection{Implementation Details}
% \paragraph{Flan-T5} We use one GPU T4 to obtain the inference results from Flan-T5 \citep{chung2022scaling} model. We use the large version from HuggingFace library (780M parameters) \citep{wolf2019huggingface}. 
% \paragraph{GPT-3} For GPT-3 \cite{brown2020language}, we use the \textit{text-davinci-003} model via the OpenAI\footnote{\url{https://platform.openai.com/docs/api-reference}} Library (\$54 USD total).

%%%%%%%%%%%%%%%%%%%%%%%%%%%%%%%%%% MACRO RESULTS %%%%%%%%%%%%%%%%%%%%%%%%%%%%%%%%%%%%%%%%%%%%%%%%%%
\begin{table}[t!]
\centering
\small
\resizebox{\columnwidth}{!}{
\begin{tabular}{|l|c|c|c|} %
\hline
\multicolumn{1}{|c|}{
\textbf{Model}}&\multicolumn{1}{|c|}{
\textbf{F1}} & \multicolumn{1}{|c|}{\textbf{P}} & \multicolumn{1}{|c|}{\textbf{R}} 
\\ \hline
Most Freq.                 & 32.44 $_{0.0}$ & 24.01 $_{0.0}$ & 50.00 $_{0.0}$ \\ \hline 
\hline
\textbf{Prompting} &&&\\
Flan-T5 (zero-shot)        & 43.90 $_{0.0}$ & 71.20 $_{0.0}$& 55.25 $_{0.0}$\\
Flan-T5 (few-shot)        & 32.91 $_{1.0}$ & 41.14 $_{0.6}$& 36.53 $_{0.4}$\\
GPT-3 (zero-shot)          & 63.65 $_{0.0}$& 65.76 $_{0.0}$& 64.20 $_{0.0}$\\
GPT-3 (few-shot)          & 69.32 $_{1.7}$&  72.12 $_{2.2}$& 70.24 $_{0.4}$\\
\hline \hline
\textbf{Image-only} &&&\\
ResNet           & 59.60 $_{0.5}$ &   59.75 $_{0.5}$   & 59.73 $_{0.5}$ \\ 
ViT              & 60.96 $_{1.2}$ &   61.62 $_{0.7}$  & 61.35 $_{0.8}$  \\ 
\hline \hline
\textbf{Text-only} &&&\\
BiLSTM-Att$^*$ \cite{zarei2020characterising}           & 66.10 $_{0.7}$ & 66.37 $_{0.7}$ & 66.27 $_{0.7}$  \\ 
BERT             & 74.35 $_{0.6}$  &  74.84 $_{0.6}$  &  74.61 $_{0.6}$ \\
BERTweet         & 76.68 $_{0.7}$ &  76.86 $_{0.5}$  & 76.76 $_{0.6}$ \\
\hline \hline
\textbf{Text \& Image} &&&\\
ViLT            & 68.44 $_{0.8}$ &  68.65 $_{0.6}$  & 68.55 $_{0.7}$  \\
LXMERT          & 66.10 $_{0.7}$  &  66.37 $_{0.7}$    & 66.27 $_{0.7}$ \\ 
MMBT            & 73.38 $_{0.6}$ &  73.89 $_{0.6}$   & 73.46 $_{0.7}$ \\
Aspect-Att$^*$ \cite{kim2021discovering}                & 75.52 $_{0.8}$ &  77.13 $_{1.1}$ & 75.80 $_{1.0}$ \\ 
ViT-BERTweet-Att (Ours)   & \textbf{77.75} $_{0.5}$  & \textbf{ 78.60} $_{0.2}$  &  \textbf{77.97} $_{0.1}$ \\
\hline
\end{tabular}
}
\caption{Macro F1-Score, precision (P) and recall (R) for commercial influencer content prediction. $^*$ denotes current state-of-the-art models for influencer commercial content detection. Subscripts denote standard deviations. Best results are in bold.}
\label{tab:results_macro_appx}
\end{table}

\begin{table}[t!]
\centering
\small
\resizebox{\columnwidth}{!}{
\begin{tabular}{|l|c|c|c|} %
\hline
\multicolumn{1}{|c|}{
\textbf{Model}}&\multicolumn{1}{|c|}{
\textbf{F1}} & \multicolumn{1}{|c|}{\textbf{P}} & \multicolumn{1}{|c|}{\textbf{R}} 
\\ \hline
Most Freq.                 & 46.04 $_{0.0}$& 42.66 $_{0.0}$ &  50.00 $_{0.0}$ \\ 
Flan-T5 (zero-shot)        & 55.43 $_{0.0}$ & 65.60 $_{0.0}$ & 54.81 $_{0.0}$ \\
Flan-T5 (few-shot)         & 40.77 $_{1.1}$ & 43.68 $_{0.4}$ & 39.52 $_{1.1}$ \\
GPT-3 (zero-shot)          & 63.96 $_{0.0}$& 63.38 $_{0.0}$& 73.64 $_{0.0}$ \\
GPT-3 (few-shot)           & 70.95 $_{0.7}$&  74.81 $_{6.4}$&  69.82 $_{4.4}$\\
BERTweet                   & 76.48 $_{1.3}$ & 74.41 $_{2.0}$ & \textbf{79.66} $_{0.4}$  \\
ViT-BERTweet-Att (Ours)    & \textbf{77.69} $_{0.1}$ & \textbf{77.41} $_{0.7}$  & 78.00 $_{0.6}$   \\
\hline
\end{tabular}
}
\caption{Macro F1-Score, precision (P) and recall (R) for commercial influencer content prediction for tweets containing text only. Subscripts denote standard deviations. Best results are in bold.}
\label{tab:res_txtonly_macro_appdx}
\end{table}

\end{document}